\documentclass[pdflatex,sn-mathphys-num]{sn-jnl}% Math and Physical Sciences Numbered Reference Style 
\usepackage{graphicx}%
\usepackage{multirow}%
\usepackage{amsmath,amssymb,amsfonts}%
\usepackage{amsthm}%
\usepackage{mathrsfs}%
\usepackage[title]{appendix}%
\usepackage{xcolor}%
\usepackage{textcomp}%
\usepackage{manyfoot}%
\usepackage{booktabs}%
\usepackage{makecell}%
\usepackage{algorithm}%
\usepackage{algorithmicx}%
\usepackage{algpseudocode}%
\usepackage{listings}%
\usepackage{subcaption}%
\usepackage{placeins}
\usepackage{float}
\usepackage{footmisc}

\usepackage{wrapfig}
\theoremstyle{thmstyleone}%
\theoremstyle{thmstyletwo}%

\theoremstyle{thmstylethree}%

\begin{document}

%\title[Article Title]{From Images to Insights: Visual Question Answering for Intelligent Plant Disease Identification}
\title[Article Title]{A Two--Stage Multitask Vision--Language Framework for Explainable Crop Disease Visual Question Answering}
%% Author details
%%=============================================================%%

\author[1]{\fnm{Md. Zahid} \sur{Hossain}}\email{zahidd16@gmail.com}
\author[2]{\fnm{Most. Sharmin Sultana} \sur{Samu}}\email{sharminsamu130@gmail.com}
\author*[1]{\fnm{Md. Rakibul} \sur{Islam}}\email{rakib.cse@aust.edu}
\author[1]{\fnm{Md. Siam} \sur{Ansary}}\email{siamansary.cse@aust.edu}

%%=============================================================%%
%% Affiliation details
%%=============================================================%%

\affil[1]{\orgdiv{Department of Computer Science and Engineering}, 
\orgname{Ahsanullah University of Science and Technology}, 
\orgaddress{\city{Dhaka}, \postcode{1208}, \country{Bangladesh}}}

\affil[2]{\orgdiv{Department of Computer Science and Engineering}, 
\orgname{BRAC University}, 
\orgaddress{\city{Dhaka}, \postcode{1212}, \country{Bangladesh}}}


%\abstract{Plant diseases cause large losses each year and challenge farmers worldwide. Early decisions need more than labels from single images. Images alone cannot explain disease stage or control needs. This work tells a different story through visual question answering. We study how questions guide models to read crop images with purpose. Existing methods ask simple questions and miss key attributes. They also suffer from bias and fragile design. Our research focuses on richer questions for plant disease understanding. The questions reflect stages, symptoms, and basic management needs. The system links visual cues with related knowledge when needed. This approach moves beyond counting and detection. It aims to support real decisions in smart agriculture. The study highlights the promise and limits of VQA for plant disease analysis.}%
\abstract{Visual question answering (VQA) for crop disease analysis requires accurate visual understanding and reliable language generation. In this work, we present a lightweight and explainable vision--language framework for crop and disease identification from leaf images. The proposed approach integrates a Swin Transformer vision encoder with sequence-to-sequence language decoders. The vision encoder is first trained in a multitask setup for both plant and disease classification, and then frozen while the text decoders are trained, forming a two-stage training strategy that enhances visual representation learning and cross-modal alignment. We evaluate the model on the large-scale Crop Disease Domain Multimodal (CDDM) dataset using both classification and natural language generation metrics. Experimental results demonstrate near-perfect recognition performance, achieving 99.94\% plant classification accuracy and 99.06\% disease classification accuracy, along with strong BLEU, ROUGE and BERTScore results. Without fine-tuning, the model further generalizes well to the external PlantVillageVQA benchmark, achieving 83.18\% micro accuracy in the VQA task. Our lightweight design outperforms larger vision--language baselines while using significantly fewer parameters. Explainability is assessed through Grad-CAM and token-level attribution, providing interpretable visual and textual evidence for predictions. Qualitative results demonstrate robust performance under diverse user-driven queries, highlighting the effectiveness of task-specific visual pretraining and the two-stage training methodology for crop disease visual question answering.\footnote{An interactive demo of the proposed Swin--T5 model is publicly available as a Gradio-based application at \url{https://huggingface.co/spaces/Zahid16/PlantDiseaseVQAwithSwinT5} for community use.}}




\maketitle

\newpage
\section{Introduction}\label{sec1}

Plant disease diagnosis plays a critical role in modern agriculture and global food security. Crops remain constantly exposed to pests, fungi and environmental stress. These factors directly affect yield and quality. Reports from the Food and Agriculture Organization of the United Nations show that crop diseases cause global losses ranging from 10\% to 30\% \cite{savary2019global} annually. Such losses threaten farm productivity and food availability. Early identification of diseases is therefore essential. Accurate and timely diagnosis can reduce damage and support effective intervention. This need has driven continuous research at the intersection of agriculture, computer vision and intelligent systems.

Despite its importance, crop disease diagnosis remains a difficult task. Farmers usually depend on agricultural experts for on-site inspection and recommendations. Experts follow a step-by-step diagnostic process. They first identify the affected plant part. They then observe visible abnormalities. Finally, they analyze disease spot characteristics such as color, shape and distribution \cite{turkouglu2018apricot}. This process requires experience, time and physical presence. Diagnostic delays allow pests and pathogens to spread quickly. Also, in many regions, expert access is limited. This makes large-scale and timely disease monitoring difficult. As a result, delayed diagnosis often leads to severe yield loss and economic damage.

To address these challenges, automated disease detection methods have been widely explored. Early computer vision approaches relied on handcrafted features and traditional classifiers \cite{turkouglu2018apricot}. These methods often required specific imaging conditions such as fixed lighting and angles \cite{mohanty2016using, ferentinos2018deep, bhuiyan2023bananasqueezenet, hossain2024deep}. Such requirements increase the deployment cost and limit real-world applications. Recent advances in deep learning have significantly improved crop disease classification accuracy. Convolutional neural networks and transformer-based models show strong performance across multiple crops \cite{arun2023effective, nandhini2022deep, vasavi2022crop, wang2023odp}. However, most of these systems operate on unimodal data, mainly images or spectral signals \cite{parez2023visual, martinelli2015advanced, zhang2024study}. They typically output only disease labels. They fail to explain symptoms, disease stages or other relevant information. This limits their practical value for decision-making and disease management.

Visual Question Answering (VQA) offers a promising direction to overcome these limitations. VQA combines image understanding with natural language processing to answer questions about images \cite{antol2015vqa, yang2016stacked}. In agriculture, VQA models try to link visual symptoms with textual queries \cite{lan2023visual, waard1993chemical, zhao2024informed}. This allows users to ask targeted questions instead of receiving fixed labels. However, existing agricultural VQA studies provide only partial insights. They often lack detailed textual descriptions of different visual attributes \cite{lu2024application}. They struggle to represent disease progression stages. They also fail to answer questions that require external knowledge, such as pathogens, control strategies or pesticide use. Current VQA benchmarks focus mainly on medical domains rather than plant pathology \cite{zhang2023pmc, he2020pathvqa, liu2021slake, lau2018dataset}. Moreover, many VQA models remain computationally heavy. This restricts their use in actual farming environments. In this context, our research asks a clear question. \textbf{``Can a lightweight Visual Question Answering framework be established for intelligent and practical plant disease identification?"}

To address these challenges, there is a clear need for efficient multimodal systems that can jointly reason over visual symptoms and textual queries while remaining computationally practical for real-world agricultural deployment. Such systems should balance accuracy, interpretability and resource efficiency to ensure usability in low-resource farming environments.

In this work, we propose a unified vision--language framework for visual question answering in plant disease analysis. The framework is designed to support plant identification, disease recognition and natural language response generation to related queries. It utilizes a two-stage training strategy to improve visual understanding while maintaining efficient inference.

The proposed approach uses a Swin Transformer-based vision encoder \cite{swin} with a text decoder for answer generation. In the first stage, the vision encoder is trained under a multitask learning framework, where plant identification and disease classification are optimized jointly through shared visual representations. This joint supervision encourages the encoder to capture both crop-level and symptom-level discriminative features, leading to more robust and semantically rich visual embeddings. In the second stage of training, the pretrained encoder is reused and kept frozen to support visual question answering. This design improves stability and reduces overall computational overhead during training.

To enable robust language generation, we integrate a transformer-based text decoder (BART \cite{bart} and T5 \cite{t5}). The decoder generates natural language answers conditioned on both visual features from vision encoder and user queries. The model exhibits robustness to diverse question formulations and open-ended queries, supporting practical real-world interaction scenarios.

We incorporate explainable AI techniques to enhance the interpretability of our proposed approach. Grad-CAM \cite{grad} visualizations are used to highlight important image regions influencing predictions. Token-level attribution is applied to each question token to analyze its contribution to the answer generation process. These analyses provide transparency and validate meaningful vision--language alignment.

Extensive experiments are conducted to thoroughly evaluate the proposed framework. The model is primarily trained and tested on the large-scale Crop Disease Domain Multimodal (CDDM) benchmark \cite{cddm} for in-domain evaluation. To further assess robustness and real-world applicability, we additionally perform cross-dataset validation on the external PlantVillageVQA dataset \cite{sakib2025plantvillagevqa} without fine-tuning, enabling evaluation under distribution shift. The quality of the generated responses is evaluated with standard natural language generation (NLG) metrics, including BLEU \cite{bleu}, ROUGE \cite{rouge} and BERTScore \cite{bertscore}. Model efficiency is also analyzed in terms of parameter count and inference latency. Finally, ablation studies investigate the impact of vision encoder pretraining and the choice of decoder on overall performance.

Our key contributions are summarized as follows:
\begin{itemize}
\item We propose a unified vision--language framework for plant and disease visual question answering using natural images.
\item We introduce a two-stage training strategy that decouples visual representation learning from language generation.
\item We demonstrate robust in-domain performance and strong cross-dataset generalization without fine-tuning. 
\item We provide comprehensive explainability analysis using Grad-CAM and token-level attribution.
\item We evaluate the framework using classification accuracy, NLG metrics and model efficiency measures.
\item We show that vision encoder pretraining significantly improves performance across all evaluation metrics.
\end{itemize}

This article is organized as follows: Section \ref{sec2} provides a summary of existing works in the literature. Section \ref{sec3} details the proposed approach. Section \ref{sec4} describes the setup used for the experiments. Section \ref{sec5} presents the results and comprehensive analysis of the results. Section \ref{sec6} discusses the limitations of this study. Finally, Section \ref{sec7} concludes the study and outlines directions for future work.

\section{Related Work}\label{sec2}
This section reviews recent research on multimodal and vision–language approaches for agricultural intelligence. It focuses on how models, data and learning strategies evolve to support accurate disease diagnosis and decision-making in agriculture.

\subsection{Visual Question Answering Frameworks for Agricultural Disease Diagnosis}
Early visual question answering systems for agricultural disease diagnosis focused on multimodal feature fusion and attention mechanisms. These systems used moderate-size datasets. The fruit tree disease decision model \cite{lan2023visual} used ResNet-152 for image features and BERT for question encoding. It applied bilinear pooling with modular co-attention. The model achieved 86.36\% accuracy on a custom orchard dataset. Attention instability and keyword misalignment reduced reliability. The wheat rust diagnostic framework \cite{nanavaty2024integrating} combined CNN classifiers with a fine-tuned BLIP vision–language model and federated learning. It achieved 97.69\% classification accuracy and a BLEU score of 0.6235. The system focused on a single crop and was sensitive to image corruption. The ILCD framework \cite{zhao2024informed} used Inception-v4, LSTM and MUTAN fusion with bias-balancing strategies. It reached 86.06\% accuracy on the CDwPK-VQA dataset. The small dataset size and weak generalization limited scalability.

Recent frameworks emphasized knowledge integration and dataset expansion. They aimed to improve reasoning depth and task diversity. The CDEK model \cite{zhao2025visual} used object detection, stacked self-attention and external knowledge from agricultural repositories and GPT-3. It achieved 89.36\% accuracy on OKiCD-VQA. It struggled with unseen diseases and real-time deployment. PlantVillageVQA \cite{sakib2025plantvillagevqa} introduced a large-scale benchmark with 193,609 expert-validated question–answer pairs. The dataset covered many crops and diseases. Models such as CLIP, LXMERT and FLAVA achieved moderate accuracy. They struggled with causal and counterfactual reasoning. The joint topic entity and intent recognition model \cite{huang2026joint} used a dual-tower multimodal Transformer with multi-task learning. It achieved up to 96.5\% accuracy for entity recognition. The framework relied only on image and text inputs.

Advanced systems extended VQA toward comprehensive agricultural decision support. These systems used multitask learning and domain knowledge graphs. The HortiVQA-PP framework \cite{li2025hortivqa} integrated segmentation-aware encoders, pest–predator modeling and knowledge-guided large language models. It achieved strong segmentation, detection and VQA performance on a diverse horticultural dataset. Regional coverage, extreme visual conditions and high computational cost remained challenges. Overall progress moved from CNN-based fusion to transformer-based and knowledge-enhanced architectures. Accuracy and reasoning capability improved across studies. Common limitations included dataset bias, limited generalization across crops and environments, lack of multimodal sensor integration and reduced robustness in real-world conditions. Future work emphasized larger datasets, zero-shot or few-shot learning, stronger attention mechanisms, deeper knowledge integration, multimodal sensing and efficient field deployment \cite{nanavaty2024integrating, zhao2024informed, zhao2025visual, lan2023visual, li2025hortivqa, huang2026joint}.

\subsection{Multimodal Deep Learning and Transformer-Based Models}
Early multimodal deep learning models for agricultural analysis focused on structured feature fusion and attention mechanisms. These models used moderate-size datasets. A transformer-based multimodal system \cite{lu2024application} integrated image, text and sensor data. It used CNN backbones, BERT, GPT and multi-head self-attention. The system reached up to 0.94 accuracy in disease detection. It performed well in captioning and object detection. The model required high computational resources. Dataset diversity was limited.
Large-scale transformer and instruction-tuned models improved multimodal reasoning through knowledge integration. Agri-LLaVA \cite{wang2024agri} used LLaVA-1.5 with large agricultural datasets. It relied on GPT-4-generated instructions. Fine-tuning improved performance by 4.87\%. The model was sensitive to rare disease classes. Computational cost remained high. BLIP-DP \cite{liang2025dynamic} focused on dynamic prompt generation guided by a VQA module. It achieved a BLEU-4 score of 83.4 on PlantVillage images. The framework relied mainly on laboratory data. Real-world robustness was limited. LLaVA-PlantDiag \cite{sharma2024llava} adapted a vision–language model for plant disease diagnosis. It used LoRA-based fine-tuning and synthetic instruction data. The model reached 96\% classification accuracy. It outperformed GPT-4 Vision. Performance depended on dataset coverage. Hallucination risks remained.
Few-shot and data-efficient multimodal frameworks addressed limited labeled data. A multimodal few-shot learning system \cite{pranith2025multimodal} used contrastive Siamese networks and prototypical classification. It included retrieval augmented generation. The system achieved 93\% accuracy on a regional Indian dataset. It generalized well to an external dataset. Synthetic data balancing was required. Caption ground truth was unavailable. Comparative analysis showed a clear shift from CNN-based fusion to transformer-based and instruction-tuned systems. Reasoning, captioning and advisory performance improved over time. Dataset bias remained common. Computational demands stayed high. Sensitivity to rare cases persisted. Real-world validation was limited. Future research emphasized larger multimodal datasets, simpler models, stronger alignment, higher robustness, richer knowledge integration, multimodal sensing and reliable deployment in diverse agricultural settings  \cite{lu2024application, wang2024agri, liang2025dynamic, pranith2025multimodal, sharma2024llava}.

\subsection{Knowledge-Enhanced and Large Language Model–Driven Agricultural Assistants}
Knowledge-enhanced and large language model–driven agricultural assistants show a clear shift toward domain-specific multimodal intelligence with conversational abilities. Agri-LLaVA \cite{wang2024agri} used a knowledge-infused LLaVA-1.5 architecture trained on over 400,000 multimodal samples. The data covered more than 221 pest and disease types. The model improved visual understanding and dialogue-based diagnosis. It struggled with rare categories and environment generalization. It required high computational resources. Future work targets hallucination reduction and deeper knowledge integration. LLaVA-PlantDiag \cite{sharma2024llava}  focused on plant pathology using LoRA fine-tuning and GPT-3.5-generated instruction data. The dataset came from PlantVillage. The model achieved 96\% classification accuracy. It outperformed GPT-4 Vision on vision–language tasks. Performance depended on synthetic data quality and dataset coverage. Future work focuses on dataset expansion and robustness. CDEK \cite{zhao2025visual} integrated explicit agricultural knowledge bases and GPT-3-generated implicit knowledge. It used fine-grained visual attention. The model achieved 89.36\% accuracy on a crop disease VQA dataset. It struggled with unseen diseases and real-time deployment. Robotic deployment was limited. Future work aims at zero-shot learning and deployment optimization. LLMI-CDP \cite{wang2025large} extended VisualGLM and ChatGLM-6B using LoRA and Q-Former alignment. The dataset included 141 disease and pest categories in Chinese. The system showed strong recognition and accurate prevention advice. Deep reasoning was limited. Inference latency was high. Generalization remained weak. Future work focuses on dataset diversity, automated labeling, efficient alignment and improved contextual reasoning.

\subsection{Datasets, Benchmarking and Task-Specific Learning Strategies}
Research on agricultural vision–language systems shows varied dataset design and task-focused learning. BLIP-DP \cite{liang2025dynamic} used a manually annotated subset of the PlantVillage dataset. It applied a fixed train–test split. The method used disease-aware dynamic prompts. The prompts came from a VQA-guided mechanism. The goal was fine-grained image captioning. PlantVillageVQA \cite{sakib2025plantvillagevqa} expanded the original PlantVillage dataset into a large VQA benchmark. The dataset included expert-verified question–answer pairs. It defined multiple cognitive task levels. It used standardized evaluation with several vision–language models. This setup revealed strengths and weaknesses in different reasoning tasks. HortiVQA-PP \cite{li2025hortivqa} built a multitask dataset from greenhouse and open-field environments. It combined segmentation, co-occurrence prediction and knowledge-guided VQA. The dataset included pest–predator annotations. It used a horticulture knowledge graph. The design supported complex decision-oriented queries. The multimodal few-shot framework \cite{pranith2025multimodal} focused on limited data and regional needs. It introduced a small dataset from Tamil Nadu. It used an external dataset for generalization testing. The framework applied contrastive pre-training, prototypical learning and retrieval-augmented querying. The goal was effective learning with few labeled samples.\\
\\
The following research gaps are identified through our extensive literature search:
\begin{itemize}
\item VQA frameworks for agricultural disease diagnosis lack standardized large-scale benchmarks that cover diverse crops, diseases and real field conditions across regions.
\item Existing datasets show strong bias toward laboratory or controlled environments, which limits cross-crop, cross-region and cross-season generalization.
\item Current task-specific learning strategies emphasize identification and description but provide weak support for causal, counterfactual and decision-oriented reasoning in VQA tasks.
\item Knowledge-enhanced and multitask datasets remain limited in multimodal diversity, with minimal integration of sensor data, temporal information and ecological context.
\item Few-shot and data-efficient learning frameworks rely heavily on synthetic augmentation and lack robust validation protocols for real-world agricultural deployment.
\end{itemize}

\section{Methodology} \label{sec3}

This section presents the proposed approach for crop disease visual question answering, including the datasets, the two-stage vision--language framework, training strategy and evaluation protocols.

\subsection{Dataset}

We use image-based crop disease datasets to train and evaluate our framework. This includes the primary CDDM dataset \cite{cddm} and the external PlantVillageVQA benchmark \cite{sakib2025plantvillagevqa} for cross-dataset evaluation. The datasets provide images paired with question–answer instances covering plant identification, disease recognition and descriptive reasoning.

\subsubsection{Crop Disease Domain Multimodal (CDDM) Dataset}
We use the Crop Disease Domain Multimodal (CDDM) dataset \cite{cddm}, which contains images of healthy and diseased crops paired with multiple question–answer (QA) instances. It covers 16 crop categories and 60 disease categories, with over one million QA pairs in total. A 90/10 QA-level split is applied for training and validation, while the default test set is used exclusively for benchmarking.

\begin{figure}[h]
  \centering
  \includegraphics[width=0.8\textwidth]{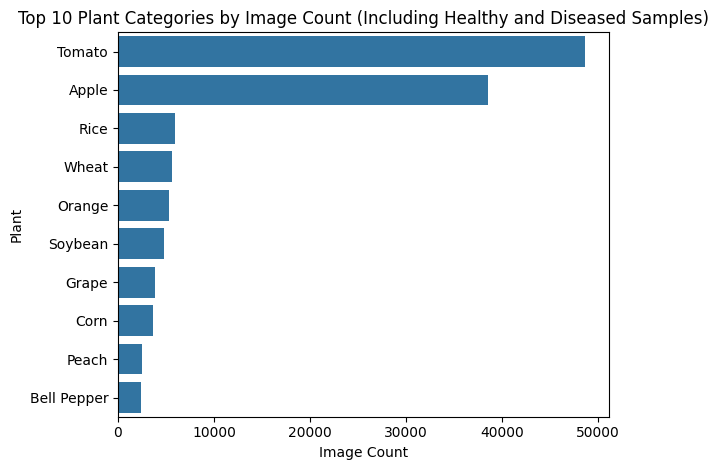}
  \caption{Distribution of plant categories by number of images in the CDDM dataset}
  \label{fig:plant_dist}
\end{figure}

\begin{figure}[h]
  \centering
  \includegraphics[width=0.8\textwidth]{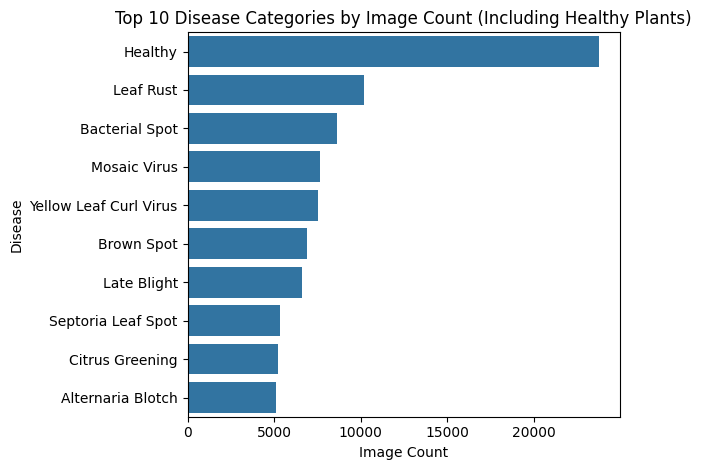}
  \caption{Distribution of disease categories by number of images in the CDDM dataset}
  \label{fig:disease_dist}
\end{figure}

\begin{figure}[h]
  \centering
  \includegraphics[width=0.8\textwidth]{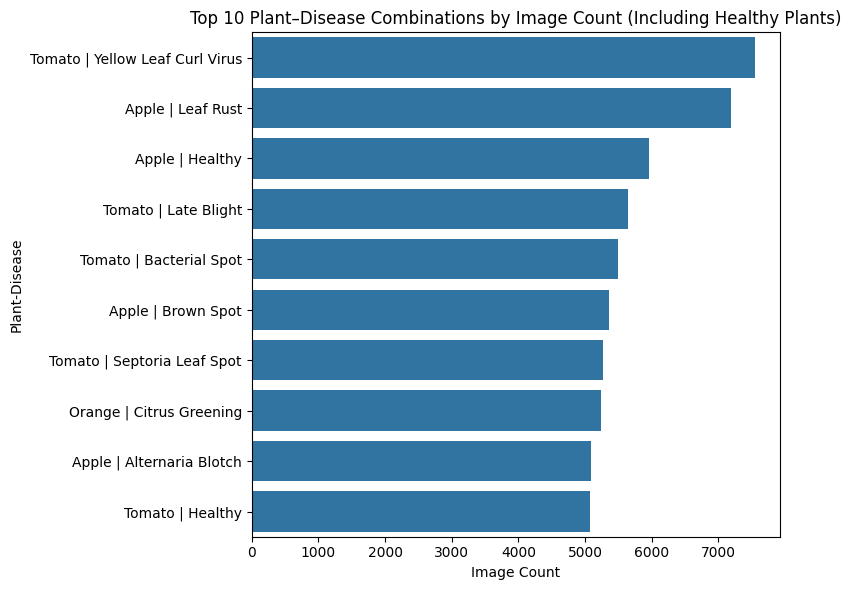}
  \caption{Distribution of plant–disease combinations by number of images in the CDDM dataset}
  \label{fig:plant_disease_dist}
\end{figure}

The average question length is 6.11 words and the average answer length is 8.92 words. The test set contains 3,963 QA pairs from 3,000 unique images and includes 292 unique answers, indicating moderate linguistic diversity. Figures~\ref{fig:plant_dist}--\ref{fig:plant_disease_dist} show the distributions of plants, diseases and plant–disease combinations.

Table~\ref{tab:dataset_summary} summarizes the sizes of the training and test splits used in this study.

\begin{table}[h!]
\caption{Sizes of the training and test splits of the CDDM dataset.}
\label{tab:dataset_summary}
\centering
\begin{tabular}{lcc}
\hline
 \textbf{Dataset Split} & \textbf{Total QA Pairs} & \textbf{Unique Images} \\
\hline
Training \& Validation & 1,056,311 & 130,150 \\
Test & 3,963 & 3,000 \\
\hline
\end{tabular}
\end{table}
\subsubsection{PlantVillageVQA Dataset}

To evaluate cross-dataset robustness and real-world generalization, we test the trained models on the external PlantVillageVQA dataset \cite{sakib2025plantvillagevqa}. The dataset contains crop leaf images paired with open-ended visual question–answer pairs spanning plant identification, disease recognition, symptom verification and descriptive reasoning.

Evaluation is conducted under a zero-shot protocol: models trained on CDDM are directly applied to PlantVillageVQA without any fine-tuning or domain adaptation. This setting measures pure cross-dataset transfer capability.

PlantVillageVQA exhibits distributional differences in both visual appearance and linguistic style, including variations in background conditions, lighting, and answer phrasing, providing a realistic out-of-distribution evaluation scenario.

Table~\ref{tab:pvqa_stats} summarizes the number of test samples per question type. Following our task focus on plant and disease recognition, three categories (visual attribute grounding, counterfactual reasoning, and existence \& sanity check) are excluded from evaluation, as they do not directly target plant or disease identification.

\begin{table}[t]
\caption{PlantVillageVQA test set statistics used for evaluation.}
\label{tab:pvqa_stats}
\centering
\begin{tabular}{lr}
\toprule
\textbf{Question Type} & \textbf{\# Samples} \\
\midrule
Plant species identification & 6,794 \\
General health assessment & 6,365 \\
Specific disease identification & 5,979 \\
Comprehensive description & 5,573 \\
Causal reasoning & 5,230 \\
Detailed verification & 2,848 \\
\midrule
\textit{Excluded categories} & \\
Visual attribute grounding & 3,205 \\
Counterfactual reasoning & 2,021 \\
Existence \& sanity check & 617 \\
\bottomrule
\end{tabular}
\end{table}
\subsection{Proposed Framework}
\begin{figure*}[t]
\centering
\includegraphics[width=0.8\textwidth]{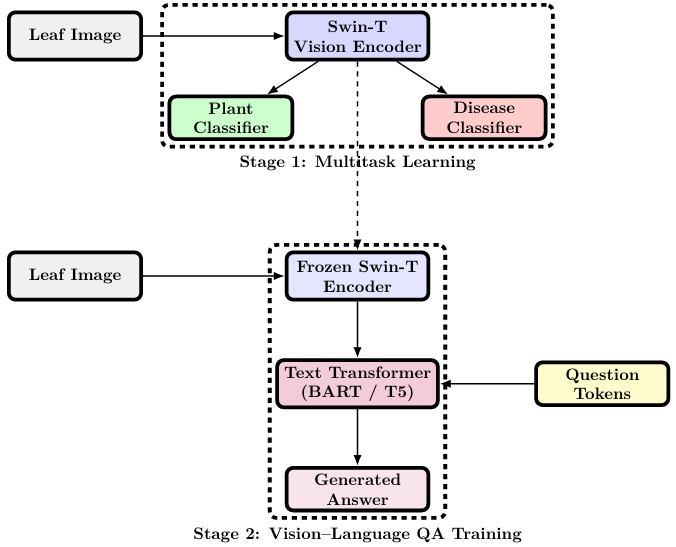}
\caption{Two-stage architecture of the proposed framework. 
Stage~1 learns plant and disease representations using a shared Swin-T encoder, while Stage~2 reuses the frozen encoder for visual question answering with a stacked text decoder.}

\label{fig:swin_qa_architecture}
\end{figure*}

The proposed framework follows a two-stage training strategy for crop disease visual question answering. The architecture is illustrated in Figure~\ref{fig:swin_qa_architecture}. The approach decouples visual representation learning from vision–language reasoning.

\textbf{Stage 1: Vision Encoder Pretraining}

In the first stage, vision encoders are trained for crop and disease classification. Two pretrained backbones are evaluated, namely CLIP ViT-B/16 \cite{clip} and Swin Transformer \cite{swin}. Both models are fine-tuned using multitask learning with shared visual features.

Given an input image, the shared backbone extracts visual representations that are fed into two task-specific classification heads corresponding to plant species and disease categories. The model is optimized using a multitask cross-entropy objective,

\begin{equation}
\mathcal{L}_{cls} = \mathcal{L}_{plant} + \mathcal{L}_{disease},
\end{equation}

which encourages the encoder to jointly capture global crop characteristics and fine-grained symptom patterns. This joint supervision produces transferable visual embeddings that are later reused for vision–language reasoning in the VQA stage.

The Swin Transformer \cite{swin} demonstrates superior classification accuracy while maintaining lower parameter complexity than CLIP ViT-B/16 \cite{clip}. Based on these results, Swin Tiny (Swin-T) \cite{swin} is selected as the vision encoder for subsequent stages.

The training hyperparameters used for fine-tuning the vision encoders are summarized in Table~\ref{tab:vit_hyp}.

\begin{table}[h!]
\caption{Hyperparameters used for training the vision encoders.}
\label{tab:vit_hyp}
\centering
\begin{tabular}{lcccc}
\hline
\textbf{Model} & \textbf{Epochs} & \textbf{Optimizer} & \textbf{Learning Rate} & \textbf{Batch Size} \\
\hline
Swin-T & 10 & AdamW & $1\times10^{-4}$ & 32 \\
ViT-B/16 & 10 & AdamW & $1\times10^{-4}$ & 32 \\
\hline
\end{tabular}
\end{table}

\textbf{Stage 2: Vision–Language Question Answering}

In the second stage, the pretrained Swin-T \cite{swin} encoder is reused for visual question answering. The encoder parameters are frozen to preserve learned visual representations. Image features are extracted as patch-level embeddings from the Swin-T backbone.

The visual embeddings are projected into the language embedding space using a learnable adapter. This projection aligns the vision features with the text decoder hidden dimension. The projected features serve as visual tokens for language conditioning.

Two decoder architectures are explored, namely BART \cite{bart} and T5 \cite{t5}. Both decoders generate natural language answers conditioned on image features and question tokens.

\textbf{Swin–BART Architecture}

For Swin–BART \cite{swin,bart}, visual embeddings are provided as encoder inputs to BART \cite{bart}. Question tokens are used as decoder inputs during training. The model is optimized using teacher forcing with cross-entropy loss.

The decoder attends to visual embeddings through standard encoder–decoder attention. This configuration supports sequence-to-sequence answer generation. The architecture is illustrated in Figure~\ref{fig:swin_qa_architecture}.

\textbf{Swin–T5 Architecture}

For Swin–T5 \cite{swin,t5}, the text decoder follows an encoder–decoder sequence-to-sequence paradigm. Visual features extracted from the frozen Swin-T encoder are used to condition answer generation through cross-modal attention.

Both global and patch-level visual features are utilized. Global representations are obtained by average pooling over patch embeddings, while patch-level features preserve fine-grained spatial information. These visual embeddings are projected to the T5 hidden dimension using a learnable multi-layer perceptron.

The projected visual features are provided as encoder inputs to T5 \cite{t5}, while question tokens are supplied to the decoder during training. Cross-attention layers within the T5 \cite{t5} decoder enable effective fusion of linguistic and visual information.

Answer generation is optimized using teacher forcing with a cross-entropy loss. Loss computation is restricted to textual output tokens, ensuring that only language generation is supervised. This design aligns with established practices in multimodal sequence-to-sequence learning and supports stable training with frozen visual encoders \cite{li2023blip,alayrac2022flamingo}.

\textbf{Training and Inference}

All vision encoders are frozen during VQA training. Only the projection layers and text decoders are optimized. Beam search with beam size 5 is applied during inference for answer generation.

The training hyperparameters used for the VQA models are summarized in Table~\ref{tab:vqa}.
\begin{table}[h!]
\caption{Hyperparameters used for training the VQA models.}
\label{tab:vqa}
\centering
\begin{tabular}{lcccc}
\hline
\textbf{Model Name} & \textbf{Epochs} & \textbf{Optimizer} & \textbf{Learning Rate} & \textbf{Batch Size} \\
\hline
Swin--BART & 2  & AdamW & $2\times10^{-5}$ & 8 \\
Swin--T5 & 3 & AdamW & $1\times10^{-4}$ & 8 \\
ViT--BART & 2  & AdamW & $2\times10^{-5}$ & 8 \\
ViT--T5 & 3 & AdamW & $1\times10^{-4}$ & 8 \\
\hline
\end{tabular}
\end{table}

\subsection{Evaluation Metrics} \label{sec:evaluation}

We evaluate the quality of the generated answers using a combination of lexical and semantic similarity metrics. These metrics assess both the correctness of key predicted entities and the overall similarity between generated and reference texts.

\begin{itemize}
  \item \textbf{Accuracy}: Accuracy measures the proportion of test samples for which the key entities (e.g., disease or condition names) are correctly identified in the generated answers. Named entities are extracted from the generated text using text extraction techniques and compared with the corresponding ground-truth annotations.

  \begin{equation}
      \text{Accuracy} = \frac{\text{Number of correctly predicted samples}}{\text{Total number of samples}}
  \end{equation}
  
  \item \textbf{BLEU (Bilingual Evaluation Understudy)}: BLEU \cite{bleu} measures the precision of $n$-gram overlap between the generated and reference texts, incorporating a brevity penalty to discourage overly short hypotheses.

  \begin{equation}
  \text{BLEU} = \text{BP} \cdot \exp\left( \sum_{n=1}^{N} w_n \log p_n \right)
  \end{equation}

  where $p_n$ denotes the modified $n$-gram precision, $w_n$ represents the weight for each $n$-gram order (typically uniform), and $\text{BP}$ is the brevity penalty defined as:
  \[
  \text{BP} =
  \begin{cases}
  1 & \text{if } c > r \\
  e^{(1 - r/c)} & \text{if } c \leq r
  \end{cases}
  \]
  Here, $c$ and $r$ denote the lengths of the candidate and reference texts, respectively.

  \item \textbf{ROUGE (Recall-Oriented Understudy for Gisting Evaluation)}: ROUGE \cite{rouge} evaluates the recall-based overlap between generated and reference texts, commonly using $n$-gram co-occurrence (ROUGE-N) or longest common subsequence measures.

  \begin{equation}
  \text{ROUGE-N} = 
  \frac{\sum_{\text{ref} \in \mathcal{R}} \sum_{\text{gram}_n \in \text{ref}} 
  \min(\text{Count}_{\text{gen}}(\text{gram}_n), \text{Count}_{\text{ref}}(\text{gram}_n))}
  {\sum_{\text{ref} \in \mathcal{R}} \sum_{\text{gram}_n \in \text{ref}} \text{Count}_{\text{ref}}(\text{gram}_n)}
  \end{equation}

  where $\text{gram}_n$ denotes an $n$-gram and counts are aggregated over all reference texts.

  \item \textbf{BERTScore}: BERTScore \cite{bertscore} measures semantic similarity by computing cosine similarities between contextualized token embeddings from a pretrained BERT model and optimally aligning tokens between generated and reference texts.

  \begin{equation}
  \text{BERTScore}_{F1} = \frac{2 \cdot P \cdot R}{P + R}
  \end{equation}

  where precision $P$ and recall $R$ are derived from token-level cosine similarity scores.
\end{itemize}

\section{Experimental Setup} \label{sec4}

The experiments were conducted using two NVIDIA T4 GPUs provided by Kaggle platform. Standard deep learning libraries such as PyTorch and Hugging Face Transformers were used. 

All images are resized to $224 \times 224$ and normalized using ImageNet \cite{deng2009imagenet} mean and standard deviation. Random seeds are fixed to ensure reproducibility across experiments.

\section{Result Analysis}\label{sec5}
We evaluate the vision–language frameworks in terms of accuracy, efficiency and interpretability. We first analyze how visual encoder pretraining improves feature quality and training stability. We then measure VQA performance on CDDM to assess in-domain effectiveness across question types. Next, we test cross-dataset generalization on PlantVillageVQA using a zero-shot setting. We also examine model complexity and inference time to estimate deployment cost. Finally, we study explainability and conduct ablation experiments to identify the contribution of each component.
\subsection{Visual Encoder Pretraining Evaluation}

Before training the vision--language VQA models, we evaluated the pretrained vision encoders on the CDDM dataset for plant and disease classification. This evaluation assesses the quality of the learned visual representations and their suitability for downstream vision--language reasoning. Accuracy was computed separately for plant identification, disease recognition and a combined metric that requires both predictions to be correct for a given sample. 

Table~\ref{tab:encoder_eval} summarizes the performance of the two models. Swin-T achieves the highest accuracy across all metrics, demonstrating both superior visual representation learning and better suitability for VQA tasks. 

\begin{figure}[!ht]
\centering
\includegraphics[width=0.8\textwidth]{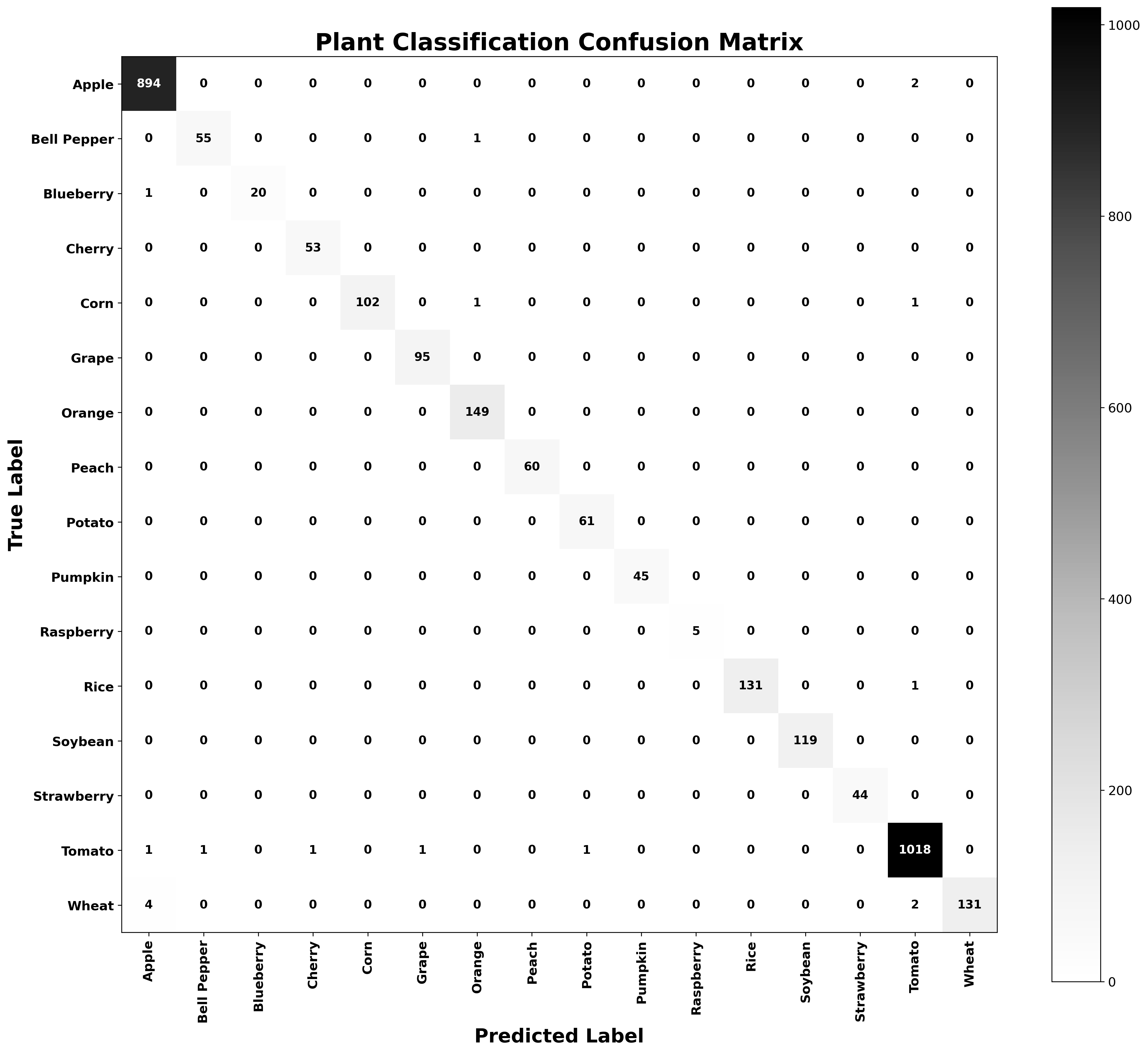}
\caption{ViT-B/16 confusion matrix for plant classification on the CDDM dataset}
\label{fig:vit_plant_confusion}
\end{figure}

\begin{figure}[!ht]
\centering
\includegraphics[width=0.8\textwidth]{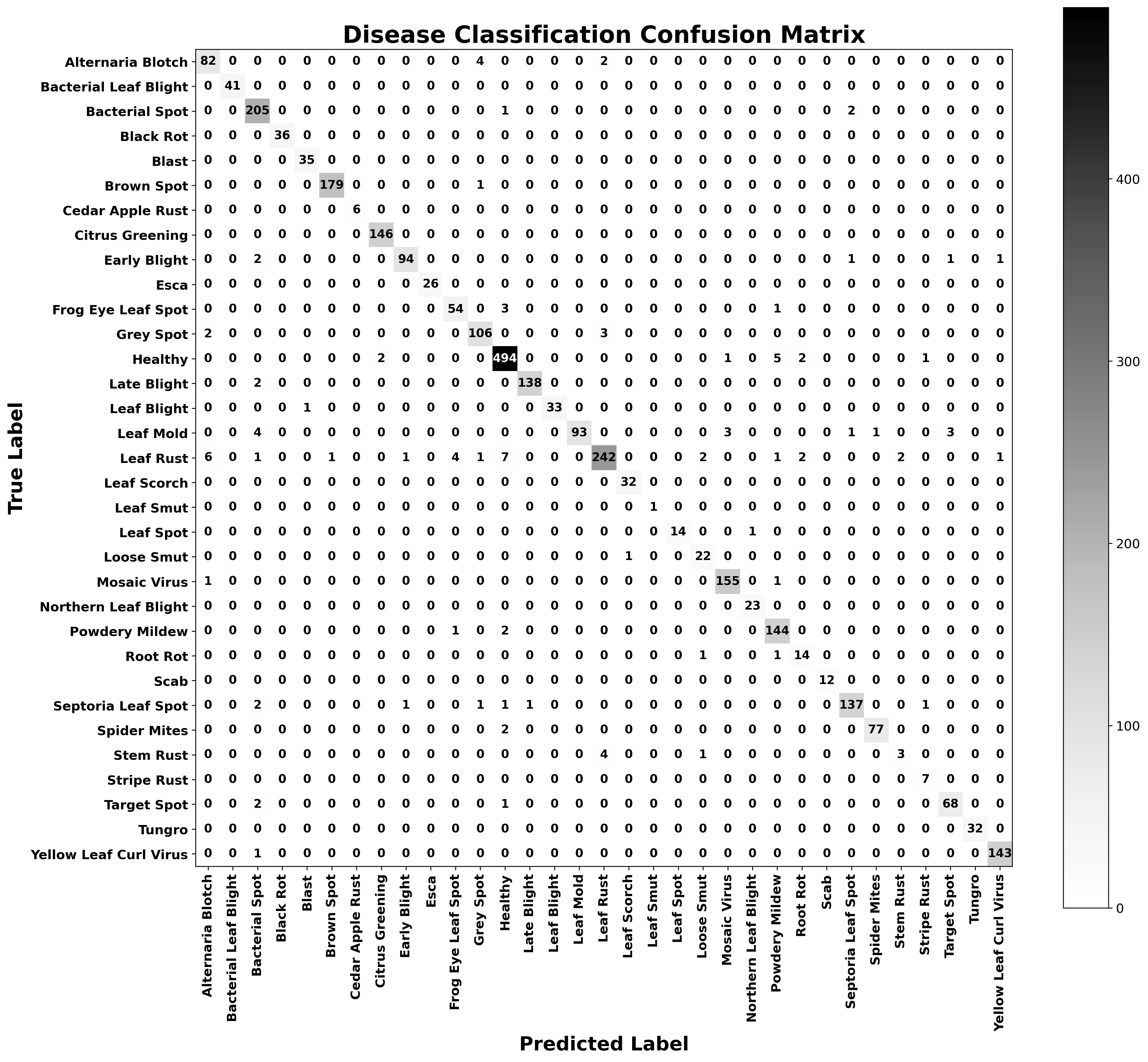}
\caption{ViT-B/16 confusion matrix for disease classification on the CDDM dataset}
\label{fig:vit_disease_confusion}
\end{figure}

\begin{figure}[!ht]
\centering
\includegraphics[width=0.8\textwidth]{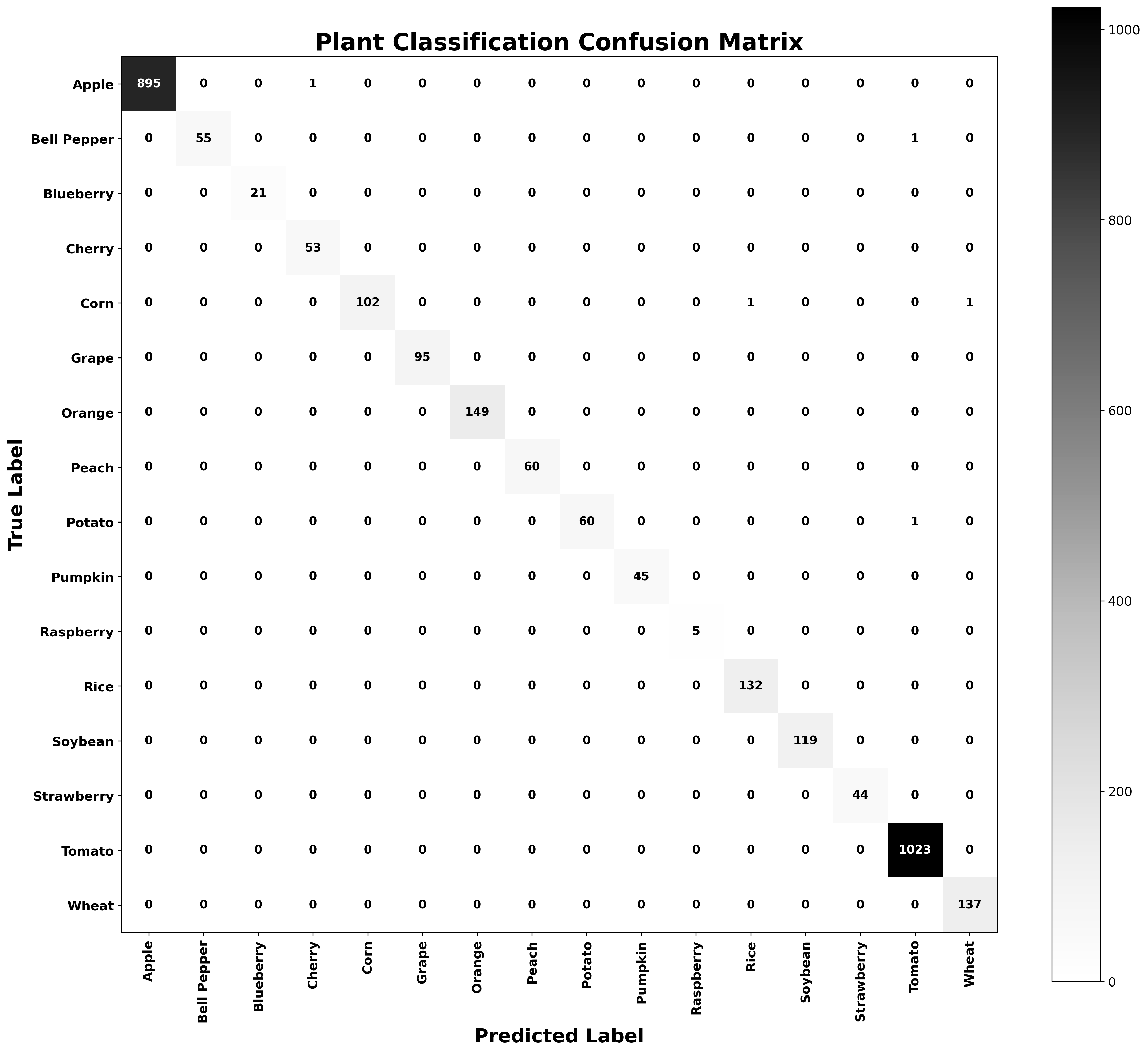}
\caption{Swin-T confusion matrix for plant classification on the CDDM dataset}
\label{fig:swin_plant_confusion}
\end{figure}

\begin{figure}[!ht]
\centering
\includegraphics[width=0.8\textwidth]{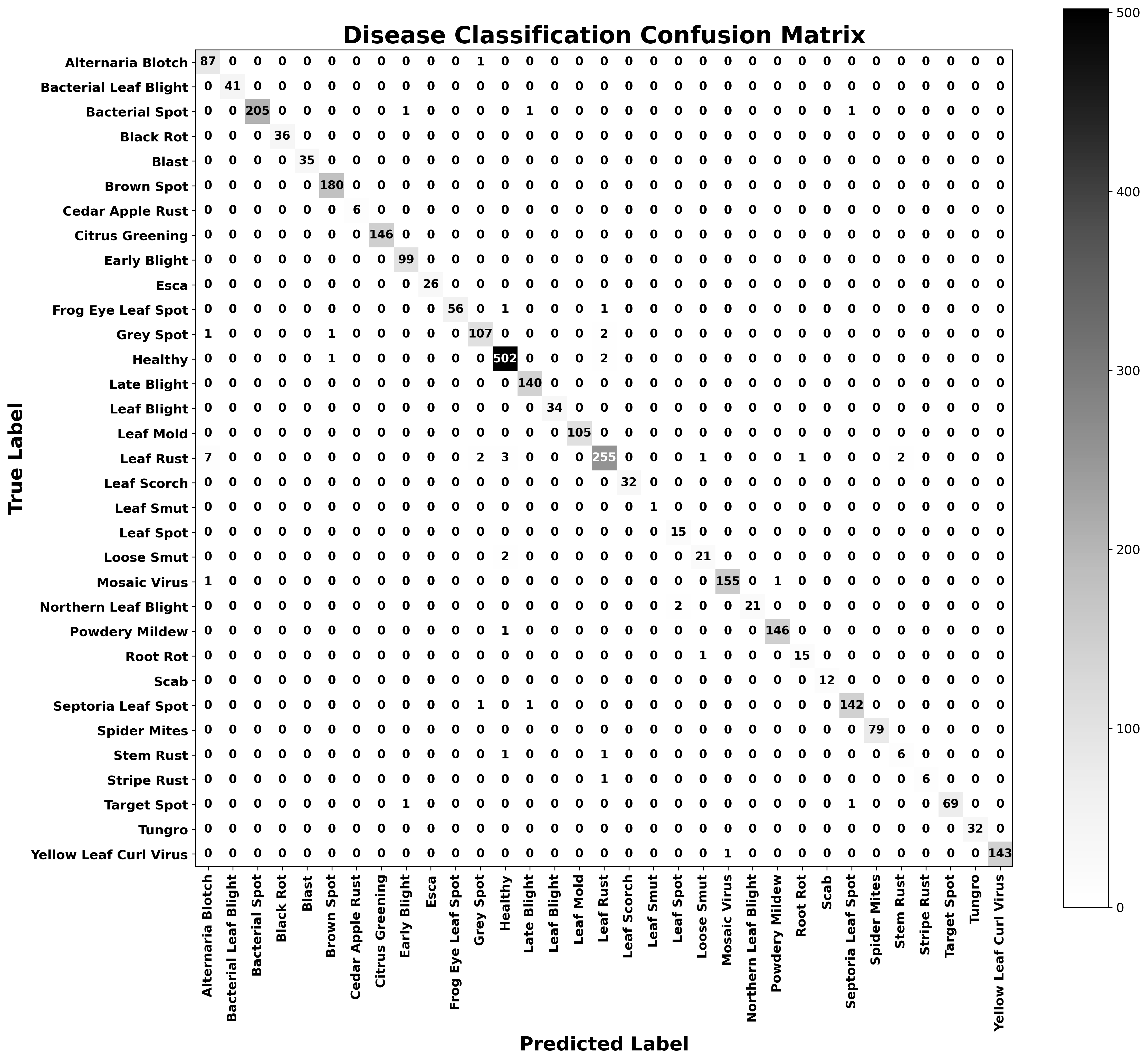}
\caption{Swin-T confusion matrix for disease classification on the CDDM dataset}
\label{fig:swin_disease_confusion}
\end{figure}

Figures~\ref{fig:vit_plant_confusion} and \ref{fig:vit_disease_confusion} show the confusion matrices for the ViT-B/16 model (plant and disease), while Figures~\ref{fig:swin_plant_confusion} and \ref{fig:swin_disease_confusion} show the corresponding matrices for Swin-T. These visualizations provide insights into per-class prediction behavior and highlight the few remaining misclassifications, which are concentrated in visually similar crop-disease combinations.

\begin{table}[h!]
\caption{Classification accuracy of different pretrained visual encoders on the CDDM dataset.}
\label{tab:encoder_eval}
\centering
\begin{tabular}{lccc}
\toprule
\textbf{Model} & \textbf{Plant Accuracy} & \textbf{Disease Accuracy} & \textbf{Combined Accuracy} \\
\midrule
ViT-B/16 & 99.40\% & 96.47\% & 96.23\% \\
Swin-T & 99.83\% & 98.50\% & 98.40\% \\
\bottomrule
\end{tabular}
\end{table}

% --- ViT-B/16 Confusion Matrices ---

% --- Swin-T Confusion Matrices ---

\subsection{VQA Performance on CDDM Dataset}
We evaluate VQA performance on the CDDM test set using multiple complementary metrics, including VQA accuracy, ROUGE, BLEU and BERTScore (F1). 
\subsubsection{Plant and Disease Identification Performance}
Table~\ref{tab:acc} compares plant and disease classification accuracy across different model configurations. The proposed Swin--T5 model achieves the highest accuracy for plant (99.94\%) and disease (99.06\%) identification. These results confirm the effectiveness of the two-stage training strategy. Models built on the Swin Transformer consistently outperform their ViT-based counterparts by a large margin in both plant and disease classification tasks, validating the choice of Swin-T as the vision encoder as discussed in the methodology. The improvement is attributed to hierarchical feature learning and strong locality modeling. These properties are well suited for fine-grained disease patterns.

Compared to LLaVA-AG \cite{cddm} and Qwen-VL-Chat-AG \cite{cddm}, the proposed Swin-based models achieve higher accuracy across both plant and disease classification. Swin--T5 attains 99.94\% plant accuracy and 99.06\% disease accuracy, outperforming Qwen-VL-Chat-AG (97.4\% and 91.5\%) and LLaVA-AG (98.0\% and 91.8\%) despite having substantially lower model complexity. Swin--BART also shows strong performance with 99.92\% plant accuracy and 97.30\% disease accuracy, indicating that accurate visual representations alone significantly boost downstream reasoning. On the other hand, ViT-based models achieve only 85–86\% plant accuracy and 84–85\% disease accuracy. Across the Swin-based variants, the T5 decoder consistently improves performance over BART, confirming the benefit of enhanced language modeling and cross-modal alignment.
\begin{table}[!ht]
\caption{Results comparison in terms of accuracy in VQA task.}
\label{tab:acc}
\centering
\begin{tabular}{lcccc}
\hline
\textbf{Model} & \textbf{Plant Identification Accuracy} &  \textbf{Disease Identification Accuracy}\\
\hline
Swin--BART (Ours)  & 99.92\%   & 97.30\% \\
Swin--T5 (Ours) & \textbf{99.94}\%&  \textbf{99.06}\%\\
ViT--BART (Ours)  & 85.87\%   & 84.68\% \\
ViT--T5 (Ours)  & 86.17\%   & 85.24\% \\
Qwen-VL-Chat-AG \cite{cddm} &  97.4\% & 91.5\%\\
LLaVA-AG \cite{cddm} & 98.0\% &  91.8\%\\
\hline
\end{tabular}
\end{table}

\subsubsection{Natural Language Generation Performance}
To evaluate answer generation quality, standard n-gram based metrics are reported. These metrics include ROUGE \cite{rouge} and BLEU \cite{bleu} scores. Table~\ref{tab:rouge_bleu} presents a comparative evaluation across different model configurations.

The proposed Swin--T5 model achieves the highest scores across all ROUGE variants and BLEU. This result indicates strong lexical overlap with ground-truth answers. Swin--BART also demonstrates high performance across all metrics. The gap between Swin--T5 and Swin--BART reflects improved language modeling capacity.

ViT-based models perform substantially worse across all metrics. This degradation aligns with their weaker visual representations. The consistent advantage of Swin-based models highlights the importance of robust visual encoding. High ROUGE-L scores further indicate improved sequence-level coherence.
\begin{table*}[ht]
 \centering
  \caption{Evaluation with N-gram based metrics.}
  \label{tab:rouge_bleu}
  \footnotesize
  \begin{tabular}{l c c c c c c}
    \toprule
    \textbf{Approach} & \makecell{\textbf{ROUGE1} \\ \textbf{F1}} & \makecell{\textbf{ROUGE2} \\ \textbf{F1}} & \makecell{\textbf{ROUGE3} \\ \textbf{F1}} & \makecell{\textbf{ROUGE4} \\ \textbf{F1}} & \makecell{\textbf{ROUGE-L} \\ \textbf{F1}} & \textbf{BLEU} \\
    \midrule
    Swin--BART &  0.9836 & 0.9786 & 0.9753 & 0.9717  & 0.9836 & 0.9727 \\
    Swin--T5 &  \textbf{0.9965}  & \textbf{0.9955} & \textbf{0.9947} & \textbf{0.9938}  & \textbf{0.9965} & \textbf{0.9940} \\
    ViT--BART &  0.8828 & 0.8799 & 0.8775 & 0.8719  & 0.8552 & 0.6320 \\
    ViT--T5 &  0.8962 & 0.8927 & 0.8875 & 0.8874  & 0.8715 & 0.6931 \\
    \bottomrule
  \end{tabular}
\end{table*}

\subsubsection{Semantic Similarity Evaluation}
Semantic consistency between generated and ground truth answers is evaluated using BERTScore F1 \cite{bertscore}. This metric captures contextual similarity beyond exact word overlap. Table~\ref{tab:bertscore} summarizes the results across model variants.

The proposed Swin--T5 model achieves the highest BERTScore F1. This result indicates strong semantic alignment with ground-truth answers. Swin--BART also attains near-perfect semantic similarity. These results reflect the effectiveness of Swin-based visual representations.

ViT-based models show noticeably lower scores. This performance gap suggests weaker cross-modal grounding. The consistent gains of T5 over BART highlight improved semantic generation.

\begin{table}[!ht]
\caption{Semantic similarity comparison using BERTScore F1.}
\label{tab:bertscore}
\centering
\begin{tabular}{p{2cm}>{\centering\arraybackslash}p{3.5cm}}
\hline
\textbf{Model} & \textbf{BERTScore F1} \\
\hline
Swin--BART   & 0.9974   \\
Swin--T5  & \textbf{0.9993}\\
ViT--BART   & 0.8843    \\
ViT--T5  & 0.8897   \\
\hline
\end{tabular}
\end{table}

\subsubsection{Model Complexity and Inference Efficiency}

We evaluate computational efficiency using model size and inference latency on a T4 GPU. Table~\ref{tab:complexity} summarizes the trade-off between performance and efficiency.

The Swin--BART model has the lowest parameter count at 167.5M. It also achieves the fastest inference time of 206.29 ms per sample. Swin--T5 increases the parameter count to 251M. This increase results in a higher inference latency of 373.35 ms.

ViT-based models exhibit higher complexity and slower inference. ViT--BART contains 226M parameters and requires 325.17 ms per sample. ViT--T5 further increases complexity to 310M parameters with 497.39 ms inference time.

Large-scale models incur substantially higher computational cost. Qwen-VL-Chat-7B \cite{cddm} requires 12.02 s per sample with 7B parameters. LLaVA-v1.5-7B \cite{cddm} reduces inference time to 9.11 s but remains significantly slower.

These large models were evaluated without fine-tuning. The reported values therefore represent approximate inference performance.

\begin{table}[h!]
\caption{Comparison of model complexity and inference efficiency.
Inference times are measured on a T4 GPU.
Results for Qwen-VL-Chat-7B and LLaVA-v1.5-7B are approximate,
as the original works \cite{cddm} report LoRA-based fine-tuning,
while we evaluate the pretrained models without fine-tuning.}
\label{tab:complexity}
\centering
\begin{tabular}{lcccc}
\hline
\textbf{Model} & \textbf{Total Parameters} &  \textbf{Average Inference Time per sample}\\
\hline
Swin--BART (Ours)  & \textbf{167.5 M}   & \textbf{206.29 ms} \\
Swin--T5 (Ours) & 251 M &  373.35 ms\\
ViT--BART (Ours)  & 226 M   & 325.17 ms \\
ViT--T5 (Ours)  & 310 M   & 497.39 ms \\
Qwen-VL-Chat-7B \cite{cddm} &  7 B & 12.02 s\\
LLaVA-v1.5-7B \cite{cddm} & 7 B &  9.11 s\\
\hline
\end{tabular}
\end{table}
 
\subsubsection{Model Explainability and Visual Reasoning Analysis}

To enhance interpretability, we employ explainable AI techniques. Grad-CAM \cite{grad} is used to identify salient image regions, and token-level attribution is applied to analyze linguistic relevance.

\begin{figure}[h]
  \centering
  \fbox{\includegraphics[width=0.8\textwidth]{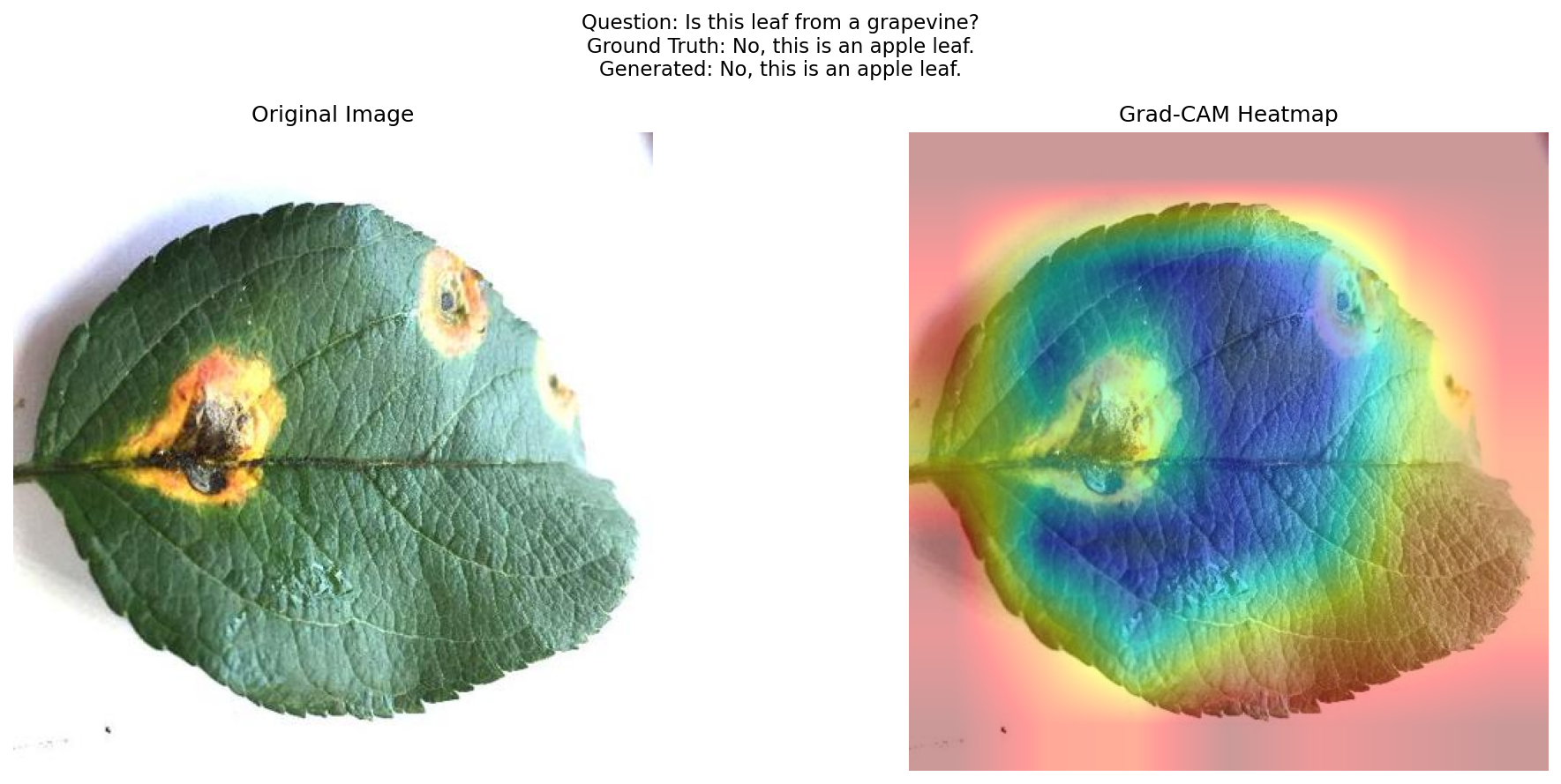}}
  \caption{Grad-CAM visualization highlighting diseased regions in an apple leaf image using Swin--T5}
  \label{fig:grad_cam1}
\end{figure}

Figure~\ref{fig:grad_cam1} presents the Grad-CAM visualization for an apple leaf image using the Swin--T5 model. The vision encoder focuses primarily on the leaf region. Increased attention is observed over the diseased areas.

\begin{figure}[H]
  \centering
  \fbox{\includegraphics[width=0.8\textwidth]{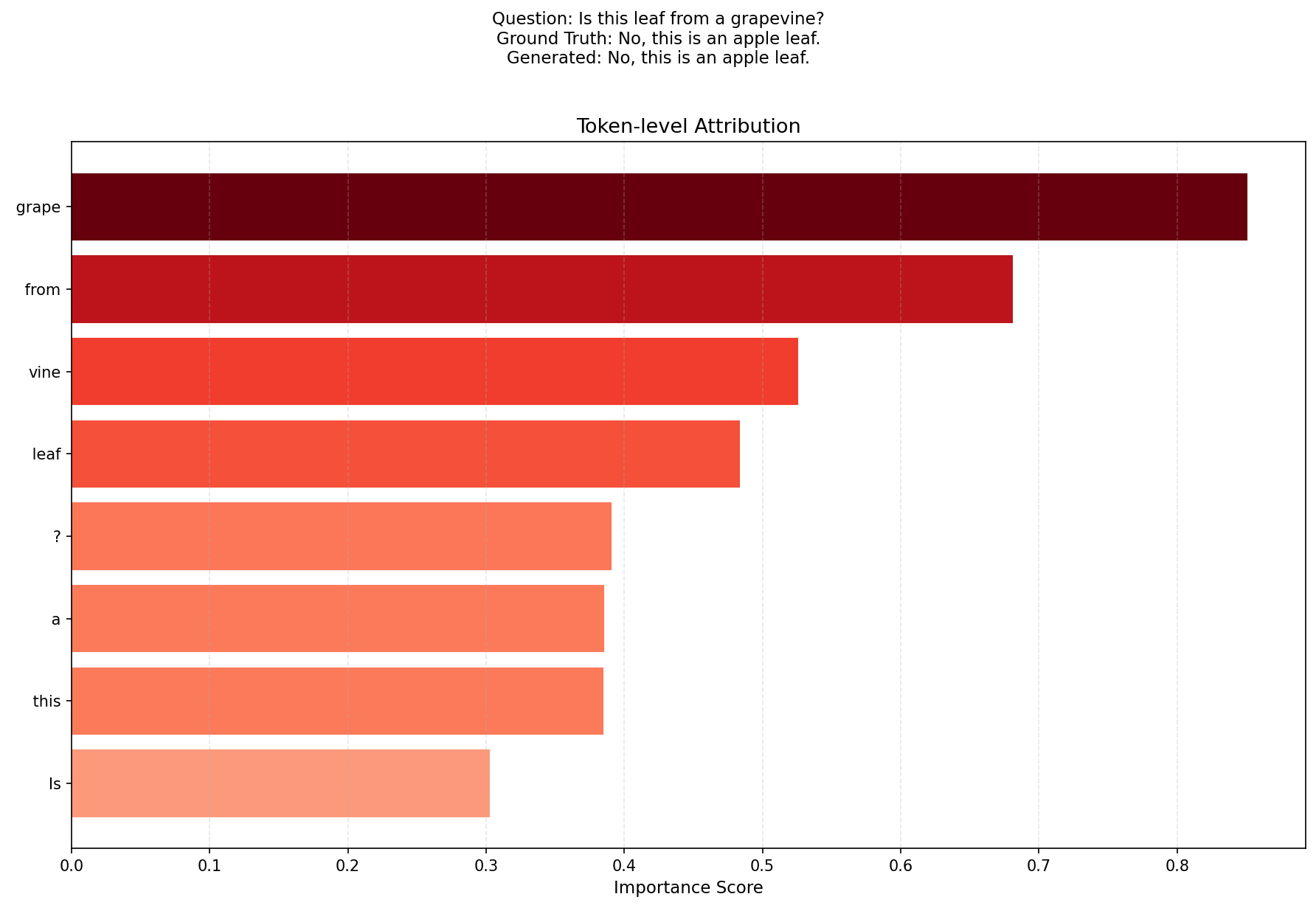}}
  \caption{Token-level attribution showing key question terms influencing answer generation}
  \label{fig:tl1}
\end{figure}
Figure~\ref{fig:tl1} illustrates token-level attribution for the corresponding question. The model assigns higher importance to keywords such as \emph{grape} and \emph{fine}. This behavior indicates effective alignment between visual and textual cues.

\begin{figure}[h]
  \centering
  \fbox{\includegraphics[width=0.8\textwidth]{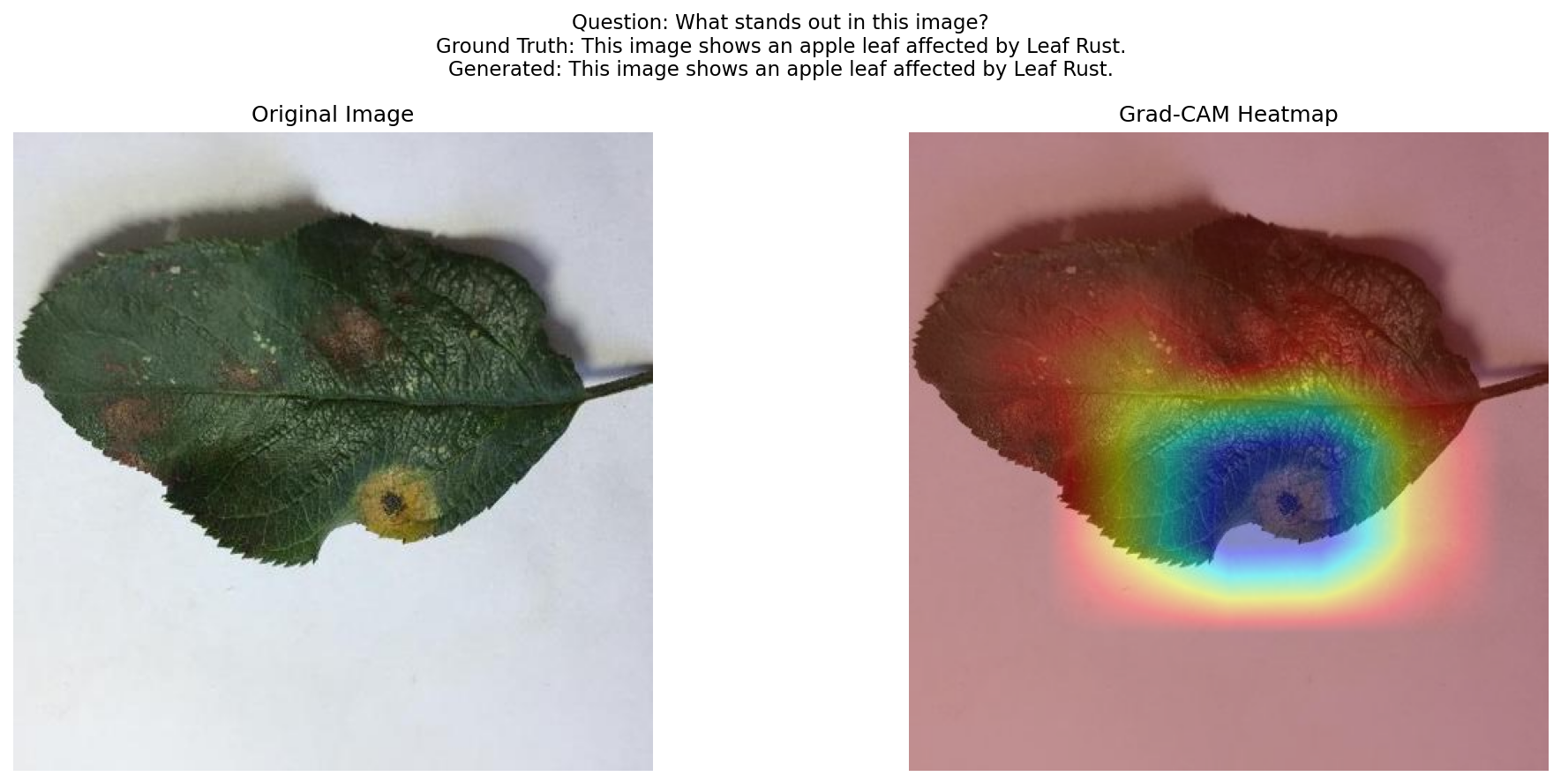}}
  \caption{Grad-CAM visualization localizing leaf rust regions in an apple leaf image}
  \label{fig:grad_cam2}
\end{figure}

\begin{figure}[H]
  \centering
  \fbox{\includegraphics[width=0.8\textwidth]{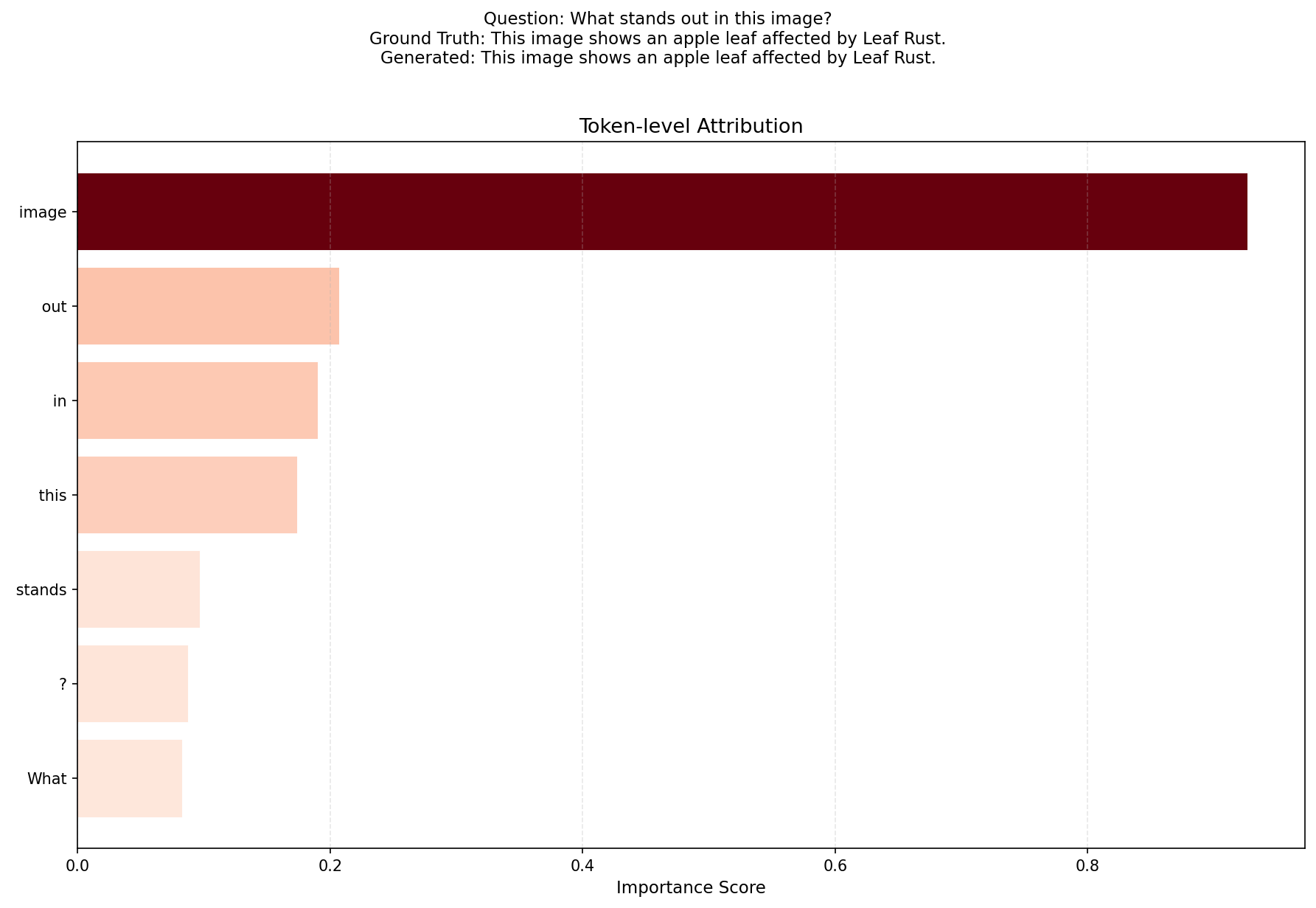}}
  \caption{Token-level attribution for an open-ended visual question emphasizing visual context}
  \label{fig:tl2}
\end{figure}

Figure~\ref{fig:grad_cam2} shows the Grad-CAM output for another apple leaf sample. The attention map highlights the region affected by leaf rust. This localization suggests disease-specific visual reasoning.
Figure~\ref{fig:tl2} presents the token-level attribution for the question \emph{“What stands out in this image?”}. The model places greater emphasis on the token \emph{image}. This reflects reliance on visual context for open-ended queries.

\begin{figure}[h]
  \centering
  \fbox{\includegraphics[width=0.8\textwidth]{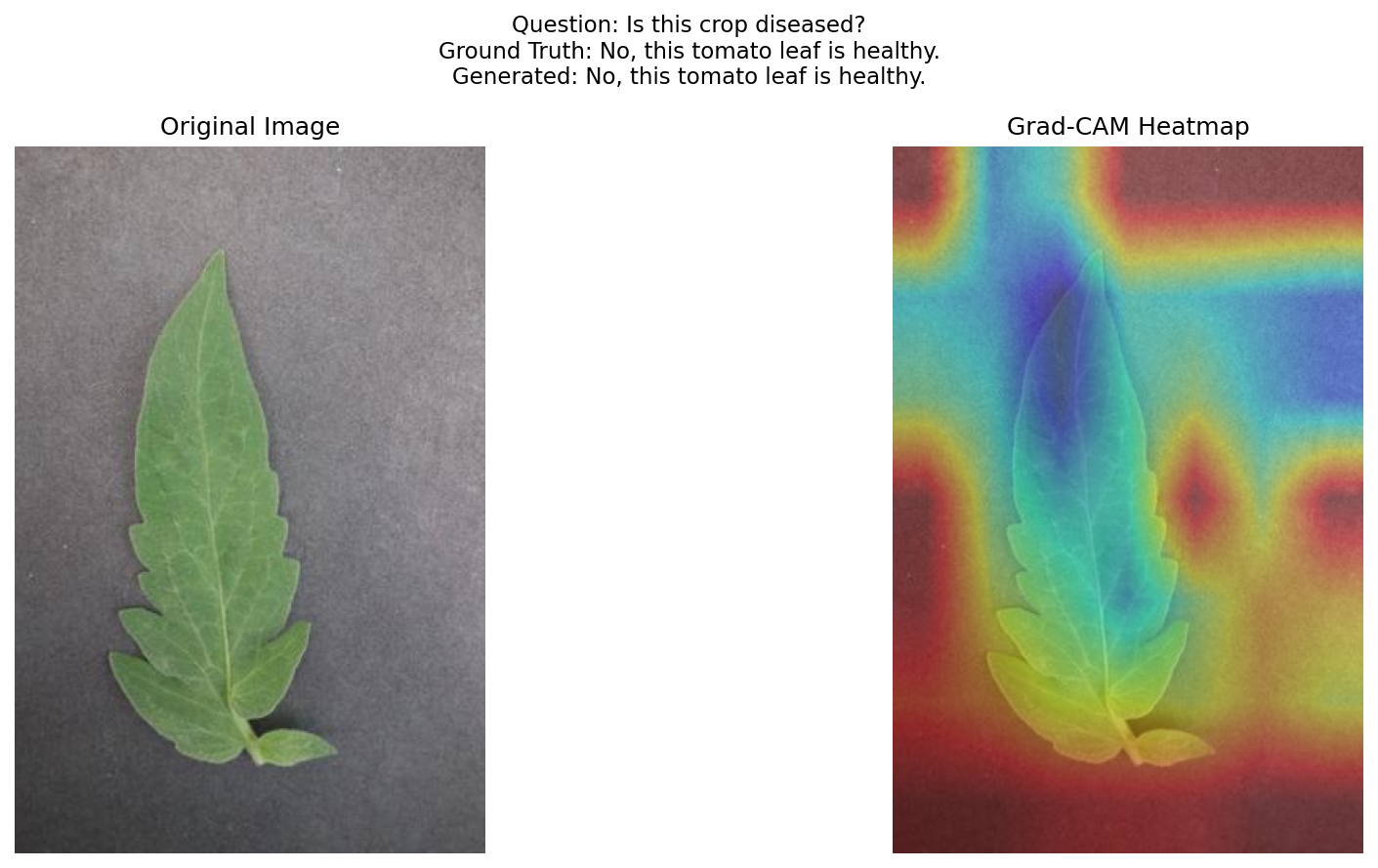}}
  \caption{Grad-CAM visualization for a healthy tomato leaf showing uniformly distributed attention}
  \label{fig:grad_cam3}
\end{figure}

\begin{figure}[H]
  \centering
  \fbox{\includegraphics[width=0.8\textwidth]{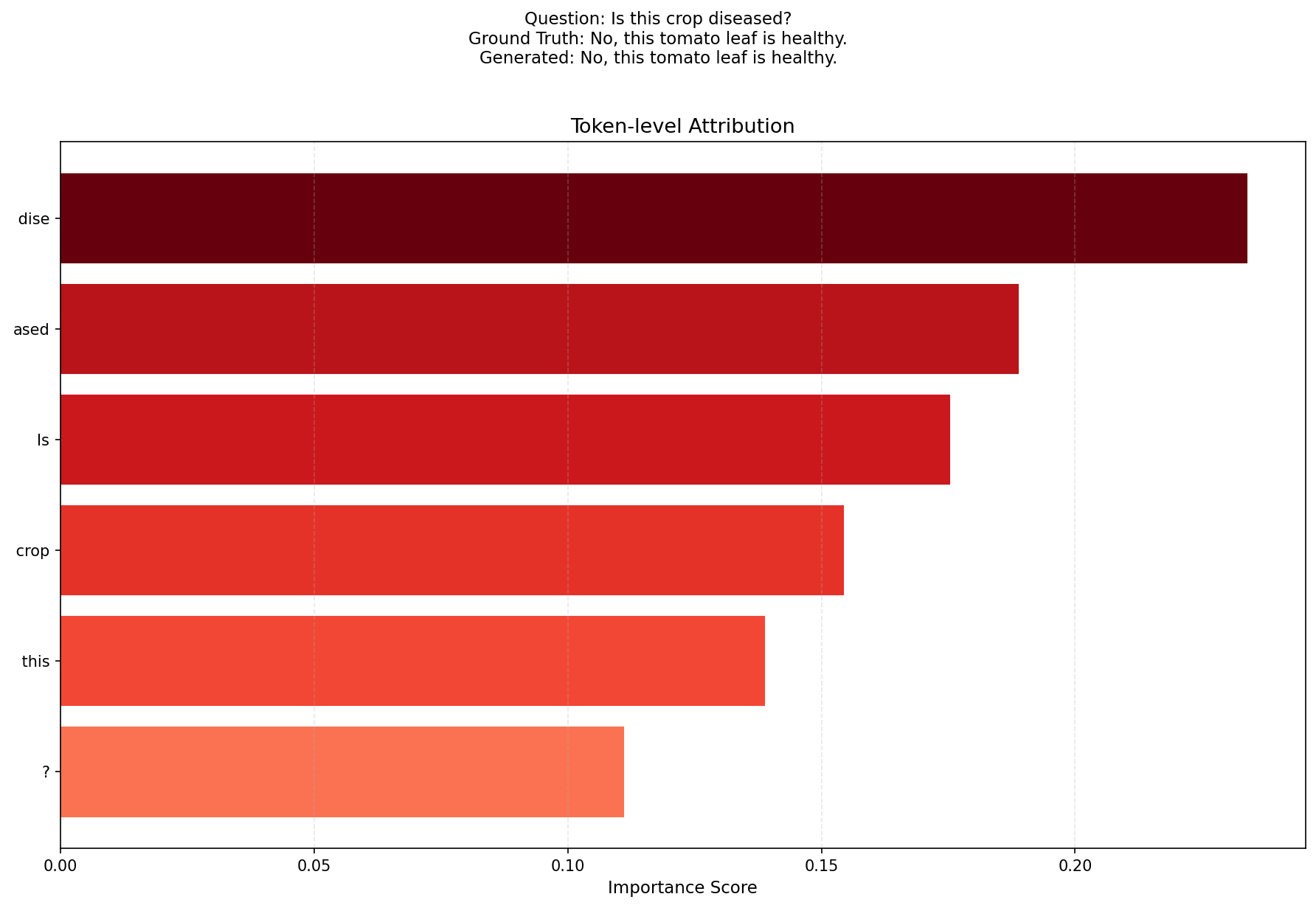}}
  \caption{Token-level attribution emphasizing diagnostic keywords in a disease identification question}
  \label{fig:tl3}
\end{figure}
Figure~\ref{fig:grad_cam3} illustrates the Grad-CAM visualization for a healthy tomato leaf. The attention is uniformly distributed across the leaf surface. No localized region dominates the activation.
Figure~\ref{fig:tl3} shows token-level attribution for the question \emph{“Is this crop diseased?”}. Higher weights are assigned to the tokens \emph{diseased} and \emph{crop}. This indicates correct sensitivity to diagnostic keywords.

Overall, the explainability results indicate coherent visual grounding. They also confirm meaningful token-level reasoning during answer generation.

\subsubsection{Qualitative Results and Robustness Analysis}

This subsection presents qualitative examples to evaluate the robustness of the proposed framework under diverse question formulations. The evaluation focuses on user-driven queries that differ from the original test questions. All qualitative results are generated using the Swin--T5 model.

\begin{figure}[h]
\centering
\fbox{\includegraphics[width=0.8\textwidth]{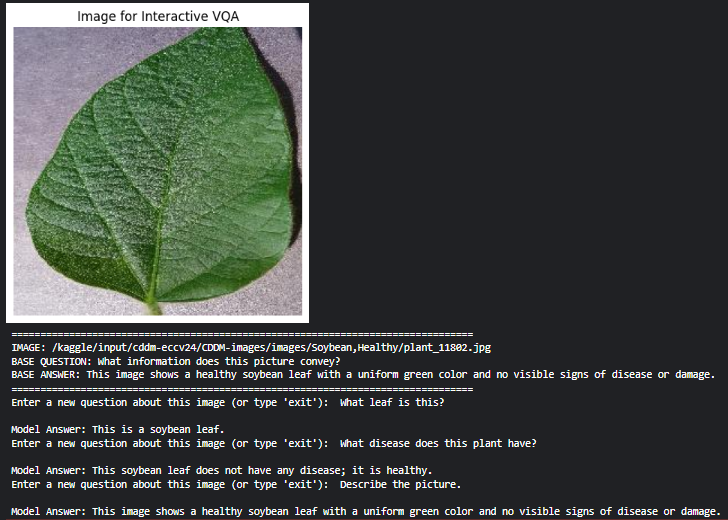}}
\caption{Qualitative example on a healthy soybean leaf with user-driven questions using Swin--T5}
\label{fig:healthy_sample}
\end{figure}
\begin{figure}[h]
\centering
\fbox{\includegraphics[width=0.8\textwidth]{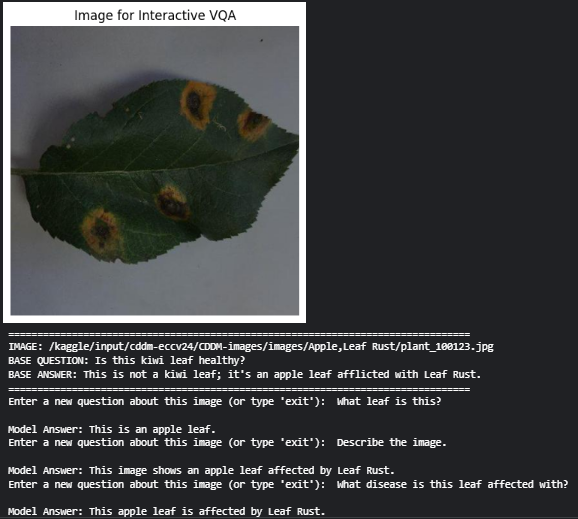}}
\caption{Qualitative example on an apple leaf affected by Leaf Rust with user-driven questions using Swin--T5}
\label{fig:diseased_sample}
\end{figure}

Figure~\ref{fig:healthy_sample} shows a healthy soybean leaf from the test set. The base question describes the visual content, and the model correctly identifies the leaf as healthy. For follow-up queries, the model consistently recognizes the crop type. It also correctly confirms the absence of disease. The responses remain semantically consistent across different question phrasings.
Figure~\ref{fig:diseased_sample} illustrates an apple leaf affected by Leaf Rust. The base question contains an incorrect plant reference, which the model successfully corrects. Subsequent user queries further validate the prediction. The model accurately identifies both the plant type and the disease. The responses remain stable across descriptive and diagnostic questions.

Overall, these examples demonstrate robustness to variations in question phrasing. The Swin--T5 model maintains correct visual grounding and semantic consistency. This behavior reflects effective vision--language alignment in interactive settings.

\subsection{Cross-Dataset Generalization on PlantVillageVQA}

To evaluate cross-dataset robustness, the trained vision--language models are directly applied to the external PlantVillageVQA benchmark using a zero-shot inference protocol, where no fine-tuning or domain adaptation is performed. All models trained on CDDM are tested as-is to assess their ability to generalize under domain shift. This section reports performance in terms of VQA accuracy, measured through entity-level correctness of plant and disease identification, along with standard NLG metrics, including BLEU, ROUGE and BERTScore, to evaluate the quality and semantic fidelity of the generated answers.
\begin{figure}[!ht]
\centering
\fbox{\includegraphics[width=0.9\textwidth]{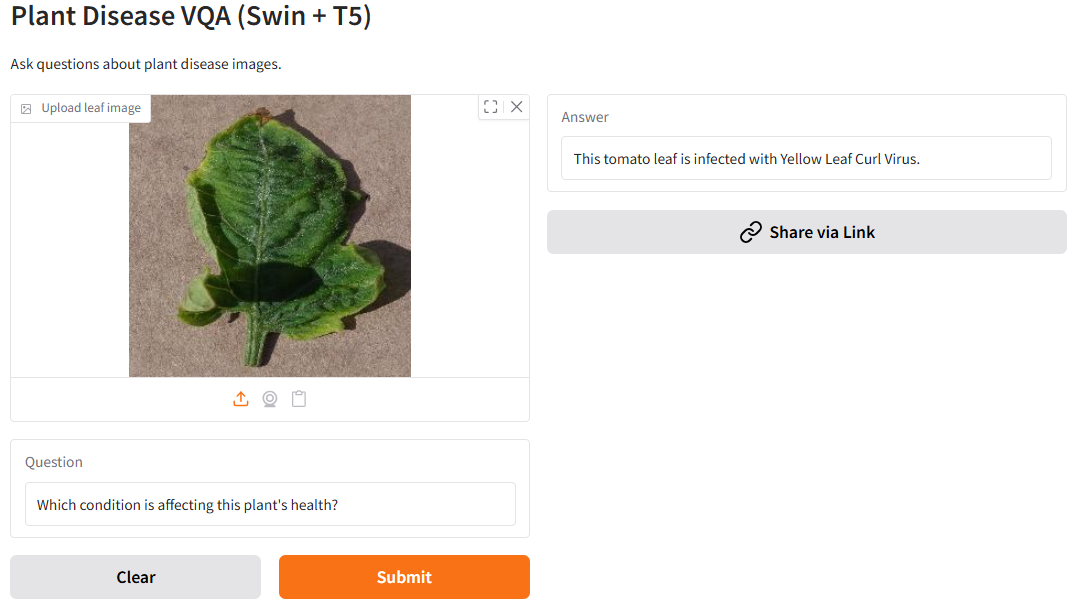}}  % Add fbox to the image
\caption{Inference result for a plant health question from the PlantVillageVQA dataset. Ground truth: ``The causal agent is the tomato yellow leaf curl virus.''}
\label{fig:plant_inference_1}
\end{figure}

\begin{figure}[!ht]
\centering
\fbox{\includegraphics[width=0.9\textwidth]{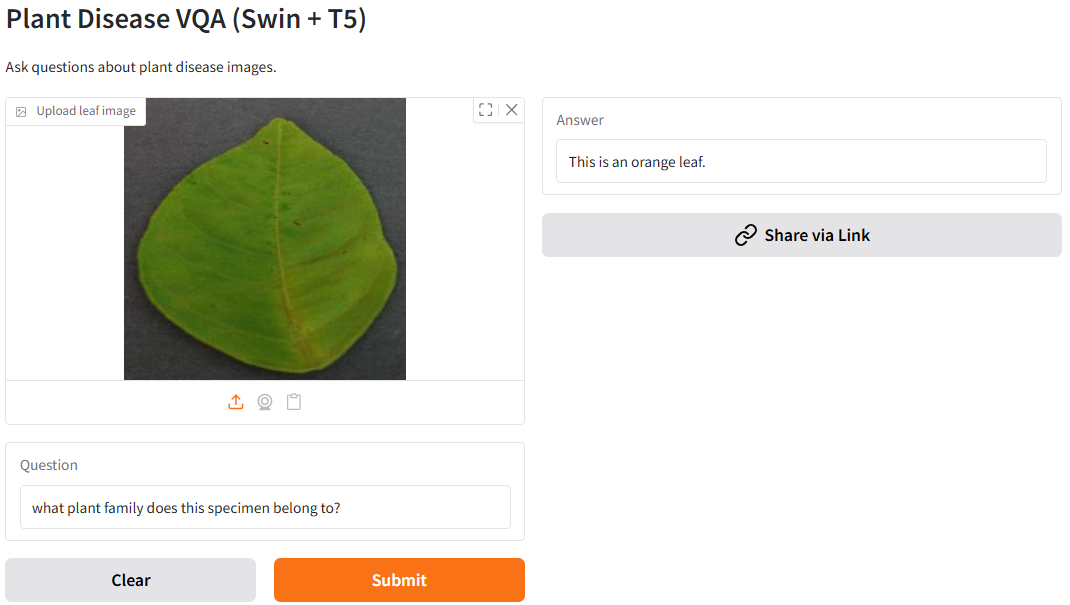}}  % Add fbox to the image
\caption{Inference result for a plant identification question from the PlantVillageVQA dataset. Ground truth: ``Orange.''}
\label{fig:plant_inference_2}
\end{figure}

As shown in Figure~\ref{fig:plant_inference_1}, the interactive application\footnote{\url{https://huggingface.co/spaces/Zahid16/PlantDiseaseVQAwithSwinT5}} enables users to input plant health-related questions, which are then processed by the Swin-T5 model. In this example, the ground truth indicates that the causal agent is the tomato yellow leaf curl virus, which the model correctly identifies.

Figure~\ref{fig:plant_inference_2} displays the interactive interface in action for a plant family identification task. The model correctly predicts that the specimen is an orange leaf, in alignment with the ground truth, demonstrating the system's ability to identify plant species in various scenarios.

\subsubsection{Quantitative Performance Evaluation}

This subsection presents the quantitative results of zero-shot evaluation on the PlantVillageVQA benchmark to assess cross-dataset generalization. All models trained on CDDM are directly applied without fine-tuning. Performance is assessed using (i) VQA accuracy for entity-level plant and disease correctness, (ii) lexical NLG metrics and (iii) semantic similarity metrics.

\paragraph{VQA Accuracy.}
Table~\ref{tab:pvqa_acc} reports micro-averaged accuracy. Swin--T5 achieves the best overall performance, indicating stronger cross-dataset transfer and better alignment between visual representations and language decoding.

\begin{table}[!ht]
\caption{Zero-shot VQA accuracy on PlantVillageVQA dataset.}
\label{tab:pvqa_acc}
\centering
\begin{tabular}{p{2cm}>{\centering\arraybackslash}p{3cm}}
\toprule
Model & VQA Accuracy (\%) \\
\midrule
Swin--BART & 79.94 \\
Swin--T5   & \textbf{83.18} \\
\bottomrule
\end{tabular}
\end{table}

\paragraph{Lexical NLG Metrics.}
Table~\ref{tab:pvqa_nlg} summarizes BLEU and ROUGE scores. Although disease and plant identification remain relatively robust, lexical overlap metrics are comparatively low, reflecting differences in answer phrasing and linguistic style between datasets.

\begin{table*}[ht]
\centering
\caption{Lexical NLG performance (BLEU and ROUGE-F1) on the PlantVillageVQA dataset.}
\label{tab:pvqa_nlg}
\footnotesize
\begin{tabular}{l c c c c c c}
\toprule
\textbf{Model} & \makecell{\textbf{ROUGE1} \\ \textbf{F1}} & \makecell{\textbf{ROUGE2} \\ \textbf{F1}} & \makecell{\textbf{ROUGE3} \\ \textbf{F1}} & \makecell{\textbf{ROUGE4} \\ \textbf{F1}} & \makecell{\textbf{ROUGE-L} \\ \textbf{F1}} & \textbf{BLEU} \\
\midrule
Swin--BART   & 0.1690 & 0.0517 & \textbf{0.0165} & \textbf{0.0078} & 0.1562 & 0.0237 \\
Swin--T5     & \textbf{0.1915} & \textbf{0.0555} & 0.0157 & 0.0066 & \textbf{0.1752} & \textbf{0.0321} \\
\bottomrule
\end{tabular}
\end{table*}

\paragraph{Semantic Similarity.}
Table~\ref{tab:pvqa_sem} reports BERTScore F1, which captures semantic consistency beyond exact word matching. Swin--T5 again demonstrates superior semantic alignment with reference answers.

\begin{table}[!ht]
\caption{Semantic similarity evaluation using BERTScore F1 on the PlantVillageVQA dataset.}
\label{tab:pvqa_sem}
\centering
\begin{tabular}{p{2cm}>{\centering\arraybackslash}p{3.5cm}}
\toprule
Model & BERTScore F1 \\
\midrule
Swin--BART & 0.1516 \\
Swin--T5   & \textbf{0.1901} \\
\bottomrule
\end{tabular}
\label{tab:pvqa_sem}
\end{table}

Overall, Swin--T5 consistently outperforms Swin--BART across accuracy, lexical overlap and semantic similarity metrics, demonstrating improved robustness under domain shift. While VQA accuracy remains relatively high in the zero-shot setting, lower BLEU and ROUGE scores suggest variations in answer wording rather than failures in visual understanding, highlighting the challenge of cross-dataset language generalization.

\subsubsection{Question-Type-wise Performance Analysis}

We further analyze robustness under domain shift by reporting question-type-wise performance on the PlantVillageVQA benchmark. Accuracy evaluates entity-level correctness of plant and disease predictions, while lexical and semantic metrics assess the quality and faithfulness of generated answers.

\paragraph{Accuracy Analysis.}
Table~\ref{tab:pvqa_qtype_acc} reports micro accuracy for each question type. 
Swin--T5 consistently outperforms Swin--BART in most reasoning-heavy and verification-oriented tasks, particularly for causal reasoning and detailed verification. 
Both models maintain strong performance for direct recognition tasks such as plant species and disease identification, indicating that visual representations transfer well across datasets.

\begin{table*}[ht]
\centering
\caption{Question-type-wise VQA accuracy (\%) on PlantVillageVQA.}
\label{tab:pvqa_qtype_acc}
\small
\begin{tabular}{lcc}
\toprule
\textbf{Question Type} & \textbf{Swin--BART (\%)} & \textbf{Swin--T5 (\%)} \\
\midrule
Causal reasoning & 62.05 & \textbf{79.94} \\
Comprehensive description & 82.04 & \textbf{83.15} \\
Detailed verification & 80.09 & \textbf{94.52} \\
General health assessment & \textbf{92.51} & 76.50 \\
Plant species identification & 81.60 & \textbf{90.55} \\
Specific disease identification & 73.77 & \textbf{79.38} \\
\bottomrule
\end{tabular}
\end{table*}

\paragraph{Lexical NLG Analysis.}
Table~\ref{tab:pvqa_qtype_nlg} summarizes BLEU and ROUGE scores. 
Lexical overlap remains modest for both models, especially for closed-form identification tasks where answers are short phrases. 
Higher scores are observed for descriptive and reasoning questions that require longer explanations.

\begin{table*}[ht] 
\centering
\caption{Lexical NLG performance (BLEU and ROUGE-F1) on the PlantVillageVQA dataset. Values are rounded to four decimal places.}
\label{tab:pvqa_qtype_nlg}
\tiny
 \setlength{\tabcolsep}{3pt}
  \begin{tabular}{l l c c c c c c}
\toprule
\textbf{Model} & \textbf{Question Type} & \makecell{\textbf{ROUGE1} \\ \textbf{F1}} & \makecell{\textbf{ROUGE2} \\ \textbf{F1}} & \makecell{\textbf{ROUGE3} \\ \textbf{F1}} & \makecell{\textbf{ROUGE4} \\ \textbf{F1}} & \makecell{\textbf{ROUGE-L} \\ \textbf{F1}} & \textbf{BLEU} \\
\midrule
\multirow{6}{*}{Swin--BART} 
& Causal reasoning & 0.2641 & 0.0963 & 0.0345 & 0.0177 & 0.2423 & 0.0373 \\
& Comprehensive description & 0.2837 & 0.1038 & 0.0309 & 0.0135 & 0.2564 & 0.0505 \\
& Detailed verification & 0.0270 & 0.0001 & 0.0001 & 0.0001 & 0.0270 & 0.0001 \\
& General health assessment & 0.0304 & 0.0001 & 0.0001 & 0.0001 & 0.0304 & 0.0001 \\
& Plant species identification & 0.0833 & 0.0001 & 0.0001 & 0.0001 & 0.0823 & 0.0001 \\
& Specific disease identification & 0.2916 & 0.1026 & 0.0316 & 0.0144 & 0.2671 & 0.0416 \\
\midrule
\multirow{6}{*}{Swin--T5} 
& Causal reasoning & 0.3241 & 0.1124 & 0.0321 & 0.0141 & 0.2937 & 0.0612 \\
& Comprehensive description & 0.2882 & 0.1002 & 0.0275 & 0.0092 & 0.2526 & 0.0488 \\
& Detailed verification & 0.0922 & 0.0001 & 0.0001 & 0.0001 & 0.0922 & 0.0001 \\
& General health assessment & 0.0228 & 0.0001 & 0.0001 & 0.0001 & 0.0228 & 0.0001 \\
& Plant species identification & 0.0954 & 0.0001 & 0.0001 & 0.0001 & 0.0939 & 0.0001 \\
& Specific disease identification & 0.3213 & 0.1124 & 0.0325 & 0.0151 & 0.2935 & 0.0584 \\
\bottomrule
\end{tabular}
\end{table*}

\paragraph{Semantic Similarity Analysis.}
Table~\ref{tab:pvqa_qtype_sem} reports BERTScore F1. 
Semantic metrics remain substantially higher than lexical scores for reasoning and descriptive questions, indicating that answers are often semantically correct despite wording differences. Swin--T5 consistently yields stronger semantic alignment.

\begin{table*}[ht]
\centering
\caption{Question-type-wise semantic similarity measured by BERTScore F1 (decimal).}
\label{tab:pvqa_qtype_sem}
\footnotesize
\begin{tabular}{lcc}
\toprule
\textbf{Question Type} & \textbf{Swin--BART} & \textbf{Swin--T5} \\
\midrule
Causal reasoning & 0.2491 & \textbf{0.3261} \\
Comprehensive description & 0.3088 & \textbf{0.3164} \\
Detailed verification & 0.0413 & \textbf{0.0710} \\
General health assessment & \textbf{0.0878} & 0.0304 \\
Plant species identification & 0.0937 & \textbf{0.1131} \\
Specific disease identification & 0.2930 & \textbf{0.3322} \\
\bottomrule
\end{tabular}
\end{table*}

Overall, these results indicate that direct recognition tasks transfer reliably across datasets, whereas free-form language generation suffers from lexical variability and stylistic differences. Nevertheless, higher semantic similarity scores suggest that many predictions remain meaning-preserving despite reduced word-level overlap, highlighting the robustness of the proposed Swin--T5 model under realistic domain shift.
\subsection{Ablation Study}

An ablation study is conducted on the CDDM dataset to examine the impact of key architectural and training components on model performance. Specifically, we analyze the effect of the training strategy by evaluating the role of vision encoder pretraining.

To assess the importance of vision pretraining, we remove the separate vision encoder pretraining stage and directly train the full VQA model by unfreezing the vision encoder. All hyperparameters are kept identical to those reported in Table~\ref{tab:vqa} to ensure a fair comparison.

Table~\ref{tab:ablation1} reports plant and disease classification accuracy under this setting. Both Swin--BART and Swin--T5 exhibit a noticeable drop in accuracy compared to their pretrained counterparts, indicating that end-to-end training without vision pretraining negatively impacts discriminative performance.

\begin{table}[h!]
\caption{Accuracy in the VQA task on the CDDM dataset when vision encoder pretraining is skipped.}
\label{tab:ablation1}
\centering
\begin{tabular}{lcc}
\hline
\textbf{Model} & \textbf{Plant Identification Accuracy} & \textbf{Disease Identification Accuracy} \\
\hline
Swin--BART & 87.16\% & 86.55\% \\
Swin--T5   & 86.63\% & 84.20\% \\
\hline
\end{tabular}
\end{table}

Beyond VQA accuracy, Table~\ref{tab:ablation2} presents results using NLG-based evaluation metrics. A consistent degradation is observed across all metrics, including ROUGE, BLEU and BERTScore, for both model variants. This confirms that skipping vision pretraining not only affects classification performance but also weakens language generation quality and vision--language alignment.

\begin{table*}[ht]
  \centering
    \caption{NLG-based evaluation metrics on the CDDM dataset when vision encoder pretraining is omitted.}
  \label{tab:ablation2}
  \footnotesize
  \setlength{\tabcolsep}{3pt}
  \begin{tabular}{l c c c c c c c}
    \toprule
    \textbf{Approach} &
    \makecell{\textbf{ROUGE1} \\ \textbf{F1}} &
    \makecell{\textbf{ROUGE2} \\ \textbf{F1}} &
    \makecell{\textbf{ROUGE3} \\ \textbf{F1}} &
    \makecell{\textbf{ROUGE4} \\ \textbf{F1}} &
    \makecell{\textbf{ROUGE-L} \\ \textbf{F1}} &
    \textbf{BLEU} &
    \makecell{\textbf{BERTScore} \\ \textbf{F1}} \\
    \midrule
    Swin--BART & 0.8931 & 0.8907 & 0.8889 & 0.8872 & 0.8930 & 0.8875 & 0.8987 \\
    Swin--T5   & 0.8882 & 0.8848 & 0.8824 & 0.8803 & 0.8879 & 0.8812 & 0.8980 \\
    \bottomrule
  \end{tabular}
\end{table*}

Overall, these results highlight the critical role of vision encoder pretraining. Removing this stage leads to consistent performance degradation across both classification and generation metrics, underscoring its importance for robust visual representation learning and effective vision--language reasoning.

\FloatBarrier
\section{Limitations}\label{sec6}

Despite strong experimental performance, the proposed framework has several limitations. The model is designed for visual understanding and question answering. It may not always provide appropriate recommendations related to disease treatment or prevention. This limitation arises from the absence of explicit agronomic knowledge.

The model also lacks broad world knowledge compared to large-scale vision--language models, such as Qwen-VL-Chat-7B and LLaVA-v1.5-7B. As a result, it may struggle with complex reasoning questions that extend beyond visual evidence. This includes queries requiring external context or expert-level explanations.

Generalization to unseen plant species remains a challenge. The model performance may degrade when evaluated on crops not present in the training data. This issue is common in supervised learning settings with limited botanical diversity.

\section{Conclusion and Future Work}\label{sec7}

This work presents a unified vision--language framework for plant and disease understanding. The model effectively integrates visual perception with natural language reasoning. Comprehensive evaluations demonstrate robustness to diverse question formulations. Explainability results provide transparency in visual and linguistic decision-making. Ablation studies confirm the importance of pretrained visual representations. Overall, the proposed approach achieves reliable and interpretable performance.

Future work will explore larger and more diverse agricultural datasets. Cross-domain generalization to unseen crops will be investigated. Multilingual question answering will be incorporated for broader accessibility. Advanced reasoning modules will be integrated to handle complex agronomic queries.

\section*{Declaration of Competing Interest}  The authors have no conflict of interest to declare relevant to this article’s content. Additionally, the authors have no relevant financial or non-financial interests to disclose. 
\section*{Data availability} The datasets used in this study is publicly available at \cite{cddm,sakib2025plantvillagevqa}.
\section*{Funding} No specific funding was received for this study.
\backmatter

\bibliography{sn-bibliography}% common bib file

\end{document}